\pdfoutput=1 

\documentclass{elsarticle}

\usepackage[T1]{fontenc}
\usepackage{lmodern}

\usepackage{lineno}

\usepackage[nolist]{acronym}

\usepackage{amssymb}
\usepackage{amsthm}
\usepackage{mathtools} 

\usepackage[labelfont=bf, figurename=Fig.]{caption} 
\usepackage{subcaption} 
\usepackage[section]{placeins} 

\usepackage{xcolor} 
\usepackage{bm} 
\usepackage[normalem]{ulem}

\usepackage{hyperref}
\usepackage[nameinlink, capitalize]{cleveref}

\usepackage{tikz}
\usepackage{pgfplots}
\pgfplotsset{compat=1.17}

\usepackage[ruled]{algorithm}
\usepackage{algorithmic}

\usepackage[%
    activate={true,nocompatibility},%
    final,tracking=true,%
    kerning=true,%
    spacing=true,%
    factor=1100,%
    stretch=10,%
shrink=10]{microtype}

\microtypecontext{spacing=nonfrench}

\newcommand{\norm}[1]{\left\lVert#1\right\rVert}    
\DeclareMathOperator{\grad}{grad}                   
\newcommand{\tr}{\mathrm{tr}\,}                     

\begin{acronym}
\acro{ARE}[ARE]{absolute relative error}
\acro{ANN}[ANN]{artificial neural network}
\acro{DIC}[DIC]{digital image correlation}
\acro{FE}[FE]{finite element}
\acro{FEM}[FEM]{finite element method}
\acro{LS-FEM}[LS-FEM]{least square finite element method}
\acro{FFNN}[FFNN]{feed-forward neural network}
\acro{MARE}[MARE]{mean absolute relative error}
\acro{PDE}[PDE]{partial differential equation}
\acro{PINN}[PINN]{physics-informed neural network}
\acro{RE}[RE]{relative error}
\acro{SEM}[SEM]{standard error of the mean}
\acrodefplural{SEM}[SEM]{standard errors of the means}
\acro{SHM}[SHM]{structural health monitoring}
\acro{VFM}[VFM]{virtual fields method}
\end{acronym}

\journal{Elsevier}

\begin{document}
\begin{frontmatter}

\title{Physics-Informed Neural Networks for Material Model Calibration from Full-Field Displacement Data}
\author{David Anton\corref{cor1}
\fnref{label1, label2}}
\ead{d.anton@tu-braunschweig.de}
\cortext[cor1]{Corresponding author}
\fntext[label2]{https://orcid.org/0000-0002-0888-0220}
\author{Henning Wessels
\fnref{label1}}
\affiliation[label1]{
    organization={Institute for Computational Modeling in Civil Engineering,
    Technische Universität Braunschweig},
    addressline={Pockelsstr. 3}, 
    city={Braunschweig},
    postcode={38106}, 
    country={Germany}}

\title{}

\begin{abstract}
The identification of material parameters occurring in constitutive models has a wide range of applications in practice. One of these applications is the monitoring and assessment of the actual condition of infrastructure buildings, as the material parameters directly reflect the resistance of the structures to external impacts. \Acp{PINN} have recently emerged as a suitable method for solving inverse problems. The advantages of this method are a straightforward inclusion of observation data. Unlike grid-based methods, such as the \ac{LS-FEM} approach, no computational grid and no interpolation of the data is required. In the current work, we propose \acp{PINN} for the calibration of constitutive models from full-field displacement and global force data in a realistic regime on the example of linear elasticity. We show that conditioning and reformulation of the optimization problem play a crucial role in real-world applications. Therefore, among others, we identify the material parameters from initial estimates and balance the individual terms in the loss function. In order to reduce the dependency of the identified material parameters on local errors in the displacement approximation, we base the identification not on the stress boundary conditions but instead on the global balance of internal and external work. We demonstrate that the enhanced \acp{PINN} are capable of identifying material parameters from both experimental one-dimensional data and synthetic full-field displacement data in a realistic regime. Since displacement data measured by, e.g., a \ac{DIC} system is noisy, we additionally investigate the robustness of the method to different levels of noise.
\end{abstract}


\end{frontmatter}


\section{Introduction}
\label{sec:introduction}
The identification of material parameters occurring in constitutive models is a major research subject in the field of solid mechanics and has a wide range of applications in practice. Probably the most obvious application is the characterization of unknown materials from experimental data. Another application that motivates this paper is continuous \ac{SHM}. Building structures and materials age during service life due to chemical and physical processes. This in turn leads to a deterioration in both reliability and quality of the structure. At the same time, external impacts to the structure, e.g., traffic loads, often increase more than assumed at design. Therefore, continuous \ac{SHM} is the prerequisite of a reliable prediction of the remaining service life of an infrastructure building \cite{chang_ReportFirstStanford_1998, entezami_StructuralHealthMonitoring_2021}. For an assessment of the actual building condition, it is crucial to consider the actual parameters of the building material. These material parameters indicate damage or material degradation, since they directly reflect the resistance of the structure to external impacts. Since we focus on steel structures in this paper, we assume linear-elastic, isotropic material behavior. The material condition of steel is thus directly reflected by Young's modulus and Poisson's ratio. Once the material parameters have been identified, they can be fed into a forward simulation, e.g., a \ac{FE} simulation, to calculate the actual resistance of the infrastructure building. Based on the simulation results, maintenance intervals and reinforcement measures can be derived.

The material parameters of interest can be identified from displacement data, e.g., measured by \acf{DIC} \cite{sutton_ImageCorrelationShape_2009}, by solving an inverse problem. The underlying equation of the inverse problem is the balance of linear momentum. Traditionally, this inverse problem is solved by numerical methods, such as the \acf{LS-FEM} approach or the \ac{VFM} \cite{avril_OverviewIdentificationMethods_2008}. Recently, it has also been shown that \acfp{PINN} \cite{raissi_PhysicsinformedNeuralNetworks_2019} are particularly suitable for solving inverse problems. \Acp{PINN} are a framework for solving forward and inverse problems involving nonlinear \acp{PDE} from the field of physics-informed machine learning \cite{karniadakis_PhysicsinformedMachineLearning_2021}. Although the idea behind this method goes back to the 1990s \cite{psichogios_HybridNeuralNetworkFirst_1992, lagaris_ArtificialNeuralNetworks_1998}, it became applicable only recently due to developments in automatic differentiation \cite{baydin_AutomaticDifferentiationMachine_2018}, software frameworks, such as TensorFlow \cite{abadi_TensorFlowLargeScaleMachine_2015} and JAX \cite{bradbury_JAXComposableTransformations_2018}, and more powerful hardware as well as the ability to record and store large amounts of data. In addition to the data, \acp{PINN} also exploit the physical laws behind the data. This can improve generalization capability and reduce the amount of required training data. Recently, \acp{PINN} have been applied to inverse problems from versatile fields in engineering and science, including the simulation of unsaturated groundwater flow \cite{depina_ApplicationPhysicsinformedNeural_2021}, super sonic flows \cite{jagtap_PhysicsinformedNeuralNetworks_2022}, tunneling \cite{xu_TransferLearningBased_2022}, biomechanics \cite{li_PhysicsInformedNeural_2021}, nano-optics \cite{chen_PhysicsinformedNeuralNetworks_2020} and epidemiology \cite{kharazmi_IdentifiabilityPredictabilityInteger_2021}, to name only a few. The main advantages of \acp{PINN} over traditional numerical methods are a straightforward inclusion of observation data and that \acp{PINN} can directly approximate the strong form of the \ac{PDE}. No computational grid is required when using \acp{PINN}.

In the context of solid mechanics, \textit{Shukla et al.} \cite{shukla_PhysicsInformedNeuralNetwork_2022} applied \acp{PINN} to quantify the microstructural properties of polycristalline nickel using ultrasound data. \textit{Rojas et al.} \cite{rojas_ParameterIdentificationDamage_2021} used \acp{PINN} in combination with some classical estimation methods to identify parameters of a damage model. For the analyses of internal structures and defects, \textit{Zhang et al.} \cite{zhang_AnalysisInternalStructures_2022} presented a general framework for identifying unknown geometric and material parameters. \textit{Haghighat et al.} \cite{haghighat_PhysicsinformedDeepLearning_2021} proposed a multi-network model for the identification of material parameters from displacement and stress data. They applied their approach to linear elasticity and further extended it to von Mises plasticity. \textit{Hamel et al.} \cite{hamel_CalibratingConstitutiveModels_2022} developed a framework for the calibration of constitutive models from full-field displacements and global force-displacement data and demonstrated it by means of hyperelastic material models. Contrary to the conventional \ac{PINN} approach, they imposed the physical constraints by using the weak form of the \ac{PDE}. This method, however, does not realize the full potential of \acp{PINN} as it relies on a computational grid for integrating the weak form of the \ac{PDE}. Furthermore, in \cite{zhang_PhysicsInformedNeuralNetworks_2020}, \textit{Zhang et al.} considered the identification of heterogeneous, incompressible, hyper-elastic material from full-field displacement data using two independent \acp{ANN}. One \ac{ANN} was used for the approximation of the displacement field and another \ac{ANN} for approximating the spatially dependent material parameter. 

In summary, the previous contributions have demonstrated that \acp{PINN} can in principle be successfully applied to inverse problems in solid mechanics and particular for material parameter identification from displacement data. However, a severe restriction is that the assumptions made in the aforementioned contributions do not match the conditions of real-world inverse problems. This mainly concerns the magnitude and quality of the measured displacements as well as the magnitude of the material parameters to be identified. In the literature, normalized domains and material parameters have been predominantly considered so far. Apart from the simplifying assumptions mentioned above, in real-world applications, we cannot assume that we know the stress state within the domain of interest. As soon as the method is applied to realistic data, problems arise that have so far been neglected in the literature.

To the best of the authors knowledge, in this paper, the linear-elastic constitutive model is the first time calibrated from full-field displacement data in a realistic regime using \acp{PINN}. The enhanced \ac{PINN} we propose in this contribution works with the strong form of the \ac{PDE} and does not rely on a computational grid and thus realizes the full potential of \acp{PINN}. In contrast, when applying the standard \ac{PINN} as proposed in \cite{raissi_PhysicsinformedNeuralNetworks_2019} without further extensions to the same displacement data, we observed that it fails in solving the inverse problem \cite{anton_IdentificationMaterialParameters_2022}. Some of the possible failure modes when training \acp{PINN} have already been pointed out in the literature. Training of \acp{PINN} is a multi-objective optimization problem where a minimum in a composite loss function with respect to data and physics is searched. Hence, a trade-off between all loss terms has to be found. The convergence issues caused by the multi-objective optimization problem are investigated in \cite{wang_UnderstandingMitigatingGradient_2021, wang_WhenWhyPINNs_2022}. In order to improve the convergence, loss term weighting \cite{wang_UnderstandingMitigatingGradient_2021} and the adaption of the neural tangent kernel \cite{wang_WhenWhyPINNs_2022} were proposed. However, as reported in \cite{henkes_PhysicsInformedNeural_2022}, we found that adaptive loss term weighting may not improve the training dynamics for problems with many different loss terms, such as those encountered in solid mechanics. The neural tangent kernel is not of much interest in practice, since it involves considerably high computational effort. Other proposed approaches to improve convergence and stability of \ac{PINN} training include adaptive activation functions \cite{jagtap_AdaptiveActivationFunctions_2020, jagtap_DeepKroneckerNeural_2022}, adaptive sampling strategies \cite{nabian_EfficientTrainingPhysicsinformed_2021} and self-adaptive \acp{PINN} using a soft attention mechanism \cite{mcclenny_SelfAdaptivePhysicsInformedNeural_2021}.

In this contribution, we further develop \ac{PINN} towards the calibration of linear-elastic constitutive models from full-field displacement and global force data in a realistic regime. We demonstrate that the standard \ac{PINN} fails in identifying material parameters from displacement data in a realistic regime due to ill-posedness of the optimization problem. Based on our observations, we therefore introduce the following extensions:
\begin{enumerate}
    \item We scale the parameters to be optimized by providing initial estimates for the material parameters and normalize both the inputs and outputs of the \ac{PINN}. We also condition the loss function by giving more weight to the data loss term. Note that the normalization and scaling are introduced in a way that does not affect the physics. 
    \item We found that when using stress boundary conditions to account for the force information, the error in the identified material parameters is strongly dependent on the local error of the displacement approximation or the data quality in the boundary region. To mitigate this dependency, we base the material parameter identification on global information and consider the balance law of internal and external work to account for the force situation.
    \item Formulating the constitutive model in terms of bulk and shear modulus leads to more robust results than trying to identify Young's modulus and Poisson's ratio directly.
\end{enumerate}
We demonstrate that the enhanced \acp{PINN} are able to identify the material parameters from both one-dimensional analytical as well as experimental and two-dimensional synthetic displacement data in a realistic regime. In addition, we conduct sensitivity analyses to show that the relative errors of the identified parameters are independent of the initial estimates in a range reasonable for real-world applications. As we approximate the integrals when calculating the internal and external work, we also perform a convergence study with respect to the number of collocation points used. Finally, we investigate the robustness of the enhanced \acp{PINN} to different levels of artificial noise imposed to the synthetic displacement data of a plate with a hole. The novelty of our contribution is thus the conditioning of the ill-posed optimization problem arising from a realistic data regime that has not been previously considered in the literature.
 
The remainder of this paper is structured as follows. In \cref{sec:preliminaries} the governing equations of solid mechanics are reviewed and a brief introduction to \acp{ANN} and particularly \acp{PINN} is provided. The further developments of the \acp{PINN} towards realistic data are presented in \cref{sec:method_development}. In \cref{sec:1D_data}, the proposed \ac{PINN} is first validated using one-dimensional analytical and experimental displacement data. Subsequently, in \cref{sec:2D_data}, the enhanced \ac{PINN} is applied to clean as well as artificially noisy, synthetic displacement data of a plate with a hole. Finally, we conclude our investigations and point out possible directions of further research in \cref{sec:conclusion}.


\section{Preliminaries} \label{sec:preliminaries} \noindent
In this section, we provide an overview of the governing equations and introduce the general notation of \acp{ANN} and \acp{PINN} which are essential for the following sections.

\subsection{Governing equations} \label{subsec:governing_equations} 
In computational mechanics, modelling usually starts with the balance of linear momentum. Assuming a static and stationary setting, the balance of linear momentum at the material point $\bm{x}$ in the domain $\Omega$ can be derived as
\begin{equation}\label{eq:balance_linear_momentum}
    \operatorname{div} \bm{\sigma}(\bm{x}) + \rho(\bm{x}) \, \bm{b}(\bm{x}) = \bm{0}, \;\;  \bm{x}\in\Omega,
\end{equation}
where $\bm{\sigma}(\bm{x})$ denotes the Cauchy stress tensor, $\rho(\bm{x})$ the density and $\bm{b}(\bm{x})$ contains accelerations related to external body forces, such as gravity. In order to close the above equation, the Cauchy stress tensor $\bm{\sigma}(\bm{x})$ is expressed as a function of strain by a constitutive model via the kinematics. 

For an isotropic, linear-elastic material, the constitutive model states
\begin{equation}\label{eq:constitutive_model_2D_E_nu}
    \bm{\sigma}(\bm{x}) = \frac{E}{1+\nu}\left(\bm{\varepsilon}(\bm{x}) + \frac{\nu}{1-2 \nu} (\tr \bm{\varepsilon}(\bm{x})) \bm{I}\right). \\
\end{equation}
The strain tensor $\bm{\varepsilon}(\bm{x})$ can be expressed by the kinematic law in terms of the primary variables, which in solid mechanics are the displacements $\bm{u}(\bm{x})$, as
\begin{equation}\label{eq:kinematic_law}
    \bm{\varepsilon}(\bm{x}) = \frac{1}{2}\left(\grad \bm{u}(\bm{x}) + \grad^T \bm{u}(\bm{x})\right).
\end{equation}
Constitutive modeling comes along with the introduction of material parameters $\bm{\kappa}$.
According to \cref{eq:constitutive_model_2D_E_nu}, the material is characterized by the Young's modulus $E$ and Poisson's ratio $\nu$, so that $\bm{\kappa}=[E, \nu]$. Under the same assumptions, in one dimension, the constitutive model in \cref{eq:constitutive_model_2D_E_nu} simplifies to
\begin{equation}\label{eq:constitutive_model_E_1D}
    \sigma(x) = E \, \varepsilon(x).
\end{equation}
with $\kappa=E$. In analogy to \cref{eq:constitutive_model_2D_E_nu}, the constitutive model can also be formulated in terms of bulk modulus $K$ and shear modulus $G$ with $\bm{\kappa}=[K, G]$, such that
\begin{equation}\label{eq:constitutive_model_2D_K_G}
    \bm{\sigma}(\bm{x}) =  K \tr {\bm{\varepsilon}}(\bm{x}) \bm{I} + 2 G \bm{\varepsilon}_{D}(\bm{x}), \\
\end{equation}
where $\bm{\varepsilon}_{D}(\bm{x}) = \bm{\varepsilon}(\bm{x}) - \tr \bm{\varepsilon}(\bm{x}) / 3 \bm{I}$ is the deviatoric part of the strain tensor. The relation between these two sets of material parameters is given by
\begin{equation}\label{eq:bulk_shear_modulus}
    K = \frac{E}{3 (1 - 2 \nu)}, \;\;
    G = \frac{E}{2 (1 + \nu)}.
\end{equation}

Substituting the stress in \cref{eq:balance_linear_momentum} by the constitutive model in \cref{eq:constitutive_model_2D_E_nu} or  \cref{eq:constitutive_model_2D_K_G}, we obtain a parameterized system of \acp{PDE}
\begin{equation}\label{eq:pde_balance_linear_momentum}
    \begin{split}
        \operatorname{div} \bm{\sigma}(\bm{x}, \bm{\kappa}) + \rho(\bm{x}) \, \bm{b}(\bm{x}) &= \bm{0}, \;\;  \bm{x}\in\Omega, \\
        \bm{\sigma}(\bm{x}_{N}) \cdot \bm{n}(\bm{x}_{N}) &= \bar{\bm{t}}(\bm{x}_{N}) \;\; \bm{x}_{N} \in \Gamma_{N} \subset \partial\Omega, \\
        \bm{u}(\bm{x}_{D}) &= \bar{\bm{u}}(\bm{x}_{D}) \;\; \bm{x}_{D} \in \Gamma_{D} \subset \partial\Omega,
    \end{split}
\end{equation}
where $\Omega$ is the domain, $\partial\Omega = \Gamma_{N} \cup \Gamma_{D}$ its boundary, $\bm{n}(\bm{x})$ the normal vector and $\bar{\bm{t}}(\bm{x})$ and $\bar{\bm{u}}(\bm{x})$ are the boundary functions for the Neumann and Dirichlet boundary condition, respectively. Finally, \cref{eq:pde_balance_linear_momentum} demonstrates that the relation between the displacements and the material parameters can be established by the balance of linear momentum from \cref{eq:balance_linear_momentum}.

In addition to the balance of linear momentum, the balance of internal and external work applies according to the principle of conservation of mechanical energy. The internal work corresponds to the strain energy and is defined as 
\begin{equation}\label{eq:internal_work}
    W_{int} = \int_{\Omega} \int_{0}^{\varepsilon} \bm{\sigma}(\bm{x}) : d\bm{\varepsilon}(\bm{x}) d\Omega,
\end{equation}
and the external work is composed of a surface and a volume part and can be calculated from
\begin{equation}\label{eq:external_work}
    W_{ext} = \int_{\partial\Omega} \int_{0}^{u} \bar{\bm{t}}(\bm{x}) \cdot  d\bm{u}(\bm{x}) d\partial\Omega + \int_{\Omega} \int_{0}^{u} \bm{b}(\bm{x}) \cdot d\bm{u}(\bm{x}) d\Omega.
\end{equation}
Assuming that the loads are static and no energy is dissipated in the form of heat, the balance of internal and external work thus states
\begin{equation}\label{eq:balance_work}
    \begin{split}
        W_{int} &= W_{ext}, \\
        \int_{\Omega} \int_{0}^{\varepsilon} \bm{\sigma}(\bm{x}) : d\bm{\varepsilon}(\bm{x}) d\Omega &= \int_{\partial\Omega} \int_{0}^{u} \bar{\bm{t}}(\bm{x}) \cdot  d\bm{u}(\bm{x}) d\partial\Omega + \int_{\Omega} \int_{0}^{u} \bm{b}(\bm{x}) \cdot d\bm{u}(\bm{x}) d\Omega.
    \end{split}
\end{equation}

\subsection{Artificial Neural Networks} \label{subsec:artificial_neural_networks}
\begin{figure}[htb]
    \centering
    \begin{tikzpicture}
        \newcommand\XO{0}
        \newcommand\YO{0}
        \newcommand\MinimumSize{0.8cm}
        \node[circle, draw, minimum size=\MinimumSize, inner sep=0pt] (I_1) at (\XO,\YO+3){$x_{1}$};
        \node[circle, draw, minimum size=\MinimumSize, inner sep=0pt] (I_2) at (\XO,\YO+1){$x_{d_{in}}$};
        \node[circle, draw, minimum size=\MinimumSize, inner sep=0pt] (H1_1) at (\XO+2.5,\YO+4){ };
        \node[circle, draw, minimum size=\MinimumSize, inner sep=0pt] (H1_2) at (\XO+2.5,\YO+2){$b^{l-1}_{k}$};
        \node[circle, draw, minimum size=\MinimumSize, inner sep=0pt] (H1_3) at (\XO+2.5,\YO+0){ };
        \node[circle, draw, minimum size=\MinimumSize, inner sep=0pt] (H2_1) at (\XO+6,\YO+4){ };
        \node[circle, draw, minimum size=\MinimumSize, inner sep=0pt] (H2_2) at (\XO+6,\YO+2){$b^{l}_{j}$};
        \node[circle, draw, minimum size=\MinimumSize, inner sep=0pt] (H2_3) at (\XO+6,\YO+0){ };
        \node[circle, draw, minimum size=\MinimumSize, inner sep=0pt] (O_1) at (\XO+8.5,\YO+3){$b^{L}_{1}$};
        \node[circle, draw, minimum size=\MinimumSize, inner sep=0pt] (O_2) at (\XO+8.5,\YO+1){$b^{L}_{d_{out}}$};
        \draw[->] (O_1) -- ++(0:1.25) node[midway, above] {$y^{L}_{1}$};
        \draw[->] (O_2) -- ++(0:1.25) node[midway, above] {$y^{L}_{d_{out}}$};
        \draw[dashed, ->] (I_1) -- (H1_1);
        \draw[dashed, ->] (I_1) -- (H1_2);
        \draw[dashed, ->] (I_1) -- (H1_3);
        \draw[dashed, ->] (I_2) -- (H1_1);
        \draw[dashed, ->] (I_2) -- (H1_2);
        \draw[dashed, ->] (I_2) -- (H1_3);
        \draw[->] (H1_1) -- (H2_1);
        \draw[->] (H1_1) -- (H2_2);
        \draw[->] (H1_1) -- (H2_3);
        \draw[->] (H1_2) -- (H2_1);
        \draw[->] (H1_2) -- (H2_2);
        \draw[->] (H1_2) -- (H2_3);
        \draw[->] (H1_3) -- (H2_1);
        \draw[->] (H1_3) -- (H2_2);
        \draw[->] (H1_3) -- (H2_3);
        \node at (\XO+4.95,\YO+2.25){$w^{l}_{jk}$};
        \draw[dashed, ->] (H2_1) -- (O_1);
        \draw[dashed, ->] (H2_1) -- (O_2);
        \draw[dashed, ->] (H2_2) -- (O_1);
        \draw[dashed, ->] (H2_2) -- (O_2);
        \draw[dashed, ->] (H2_3) -- (O_1);
        \draw[dashed, ->] (H2_3) -- (O_2);
        \draw[dotted] (\XO,\YO+1.75) -- (\XO,\YO+2.25);
        \draw[dotted] (\XO+2.5,\YO+2.75) -- (\XO+2.5,\YO+3.25);
        \draw[dotted] (\XO+2.5,\YO+0.75) -- (\XO+2.5,\YO+1.25);
        \draw[dotted] (\XO+6,\YO+2.75) -- (\XO+6,\YO+3.25);
        \draw[dotted] (\XO+6,\YO+0.75) -- (\XO+6,\YO+1.25);
        \draw[dotted] (\XO+8.5,\YO+1.75) -- (\XO+8.5,\YO+2.25);
        \node at (\XO,\YO+5){Layer $0$};
        \node at (\XO+2.5,\YO+5){Layer $l-1$};
        \node at (\XO+6,\YO+5){Layer $l$};
        \node at (\XO+8.5,\YO+5){Layer $L$};
    \end{tikzpicture}
    \caption{Schematic representation of a fully connected \acf{FFNN} according to \cite{berg_UnifiedDeepArtificial_2018}.}
    \label{fig:schema_ANN}
\end{figure}
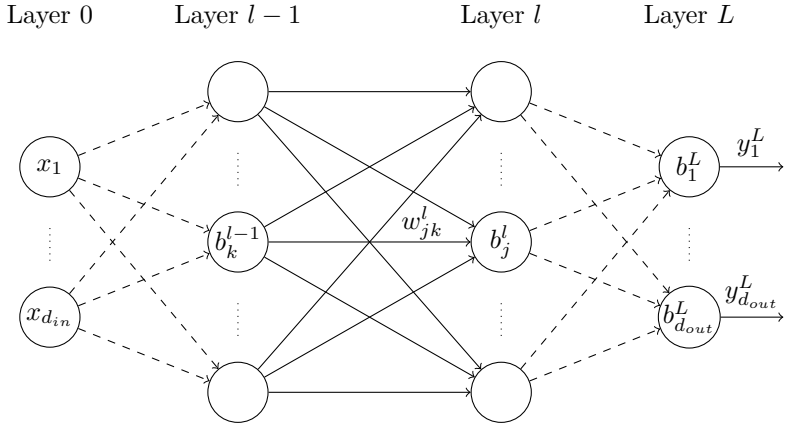

\Acfp{ANN} are global, smooth function approximators defining a mapping $\mathbb{R}^{d_{in}} \rightarrow \mathbb{R}^{d_{out}}$ from an input space to an output space. The computational units of an \ac{ANN} are called neurons and are typically arranged in an input, an output and any number of hidden layers. In the following, we consider fully connected \acp{FFNN} with $L+1$ layers. These \acp{FFNN} each have in total $L-1$ hidden layers, where layer $0$ is the input layer and layer $L$ the output layer. In a fully connected \ac{FFNN}, the neurons of each two successive layers are connected. The weight of the connection between neuron $k$ in layer $l-1$ and neuron $j$ in layer $l$ is denoted by $w^{l}_{jk}$. All weights between layer $l-1$ and $l$ can then be combined in the weight matrix $\bm{W}^{l}$, whose entries are $w^{l}_{jk}$. In addition, the neurons in the hidden layers and the output layer each have a bias, where $b^{l}_{j}$ denotes the bias of neuron $j$ in layer $l$. Similarly, the biases of all neurons in layer $l$ can be combined in vector $\bm{b}^{l}$ with entries $b^{l}_{j}$. Thus, the weights and biases of all neurons are the trainable parameters of the \ac{ANN} $\bm{\theta}=\{\bm{W}^{l},\bm{b}^{l}\}_{1\leq l \leq L}$. The schematic structure of an \ac{ANN} is illustrated in \cref{fig:schema_ANN}. In the forward pass, the output of the neurons in the hidden layers and the output layer are computed from the sum of their weighted inputs and their bias as an argument of an activation function. 

The mapping from an input to an output vector by a \ac{FFNN} can be formulated recursively according to \cite{berg_UnifiedDeepArtificial_2018}. The weighted input $z^{l}_{j}$ of neuron $j$ in  the hidden layer or output layer $l$ with an upstream layer consisting of $N_{l-1}$ neurons is defined as
\begin{equation}\label{eq:ann_weighted_input}
    z^{l}_{j}=\sum_{k=1}^{N_{l-1}}w^{l}_{jk}y^{l-1}_{k}+b^{l}_{j}, \; \forall \, l \in [1, L].
\end{equation}
In \cref{eq:ann_weighted_input}, $y^{l-1}_{k}$ is the output of neuron $k$ in the upstream layer $l-1$ given by
\begin{equation}\label{eq:ann_output}
    y^{l-1}_{k}=\phi^{l-1}_{k}(z^{l-1}_{k}).
\end{equation}
Here, $\phi^{l-1}_{k}$ is the activation function of neuron $k$ in layer $l-1$. Throughout this paper, for all neurons in a layer the same activation function is used, denoted by $\phi^{l}$ for layer $l$. Inserting \cref{eq:ann_output} in \cref{eq:ann_weighted_input} we obtain in symbolic notation
\begin{equation}\label{eq:ann_weighted_input_symbolic}
    \bm{z}^{l}=\bm{W}^{l}\bm{\phi}^{l-1}(\bm{z}^{l-1})+\bm{b}^{l}, \; \forall \, l \in [1, L],
\end{equation}
where $\bm{z}^{l-1}$ contains the weighted inputs of all neurons in layer $l-1$ and $\phi^{l-1}$ is applied elementwise. In the hidden layers, nonlinear functions, such as hyperbolic tangent, are usually used as activation functions. The activation of the output neurons is typically computed by the identity function. Since we do not use any activation function in the input layer, for the input layer $l=0$ applies
\begin{equation}\label{eq:ann_input}
    \bm{\phi}^{0}(\bm{z}^{0})=\bm{x},
\end{equation}
where $\bm{x}$ is the input vector to the \ac{ANN}. Given \cref{eq:ann_weighted_input,eq:ann_output,eq:ann_weighted_input_symbolic,eq:ann_input}, the output vector $\bm{y}^{L}$ of the \ac{FFNN} can be defined recursively as a function of $\bm{x}$ as follows:
\begin{equation}\label{eq:ann_definition}
    \begin{split}
    &\bm{y}^{L}=\bm{\phi}^{L}(\bm{z}^{L}) \\
    &\bm{z}^{L}=\bm{W}^{L}\bm{\phi}^{L-1}(\bm{z}^{L-1})+\bm{b}^{L} \\
    &\bm{z}^{L-1}=\bm{W}^{L-1}\bm{\phi}^{L-2}(\bm{z}^{L-2})+\bm{b}^{L-1}\\
    &\vdots \\
    &\bm{z}^{2}=\bm{W}^{2}\bm{\phi}^{1}(\bm{z}^{1})+\bm{b}^{2} \\
    &\bm{z}^{1}=\bm{W}^{1}\bm{x}+\bm{b}^{1}.
    \end{split}
\end{equation}
The definition in \cref{eq:ann_definition} demonstrates that \acp{ANN} are highly parameterized, nonlinear, composed functions.

The training of \acp{ANN} is an optimization problem. The objective of the optimization problem is the loss function $\mathcal{L}(\bm{\theta}; \bm{\mathcal{T}})$ which depends both on the trainable \ac{ANN} parameters $\bm{\theta}$ and the training data $\bm{\mathcal{T}}$. For the optimization of the \ac{ANN} parameters $\bm{\theta}$, usually gradient based optimization algorithms are used. The required gradient of the loss function with respect to the \ac{ANN} parameters $\bm{\theta}$ can be calculated using automatic differentiation \cite{baydin_AutomaticDifferentiationMachine_2018}.

It is well known that \acp{ANN} are universal function approximators \cite{cybenkoApproximationSuperpositionsSigmoidal1989, hornik_MultilayerFeedforwardNetworks_1989, li_SimultaneousApproximationsMultivariate_1996}. Provided an \ac{ANN} has a sufficient number of parameters, then according to the universal approximation theorem the \ac{ANN} can theoretically approximate any continuous function and its derivatives to an arbitrarily small error with mild assumptions on the activation function only. It should be noted, however, that the issue of optimal training of \acp{ANN} has not yet been solved. For a more in-depth introduction to deep learning, the reader is referred to standard text books, e.g., \cite{goodfellow_DeepLearning_2016}.

\subsection{Physics-Informed Neural Networks} \label{subsec:physics_informed_neural_networks}
\Acfp{PINN} are a deep learning framework for solving forward and inverse problems involving \acp{PDE} recently emerged in the field of scientific machine learning \cite{raissi_PhysicsinformedNeuralNetworks_2019}. The key characteristic of \acp{PINN} is the design of the loss function that enables leveraging physical knowledge during the training process. This succeeds by directly including the governing \acp{PDE} into the loss function as regularizing terms. The \ac{PINN} acts as a global function approximator of the hidden solution of the \ac{PDE}. Due to the regularizing terms in the loss function, the solution space is constrained and the \ac{ANN} is enforced to satisfy the governing \acp{PDE}. For a more detailed overview of scientific machine learning, we refer to \cite{karniadakis_PhysicsinformedMachineLearning_2021}.

Although \acp{PINN} can be applied to more general problems \cite{karniadakis_PhysicsinformedMachineLearning_2021}, in the course of this paper we consider stationary elliptic, nonlinear \acp{PDE}. Following the notation from \cite{moseley_FiniteBasisPhysicsInformed_2021}, these \acp{PDE} can be formulated as
\begin{equation}\label{eq:pinn_pde_bcs}
    \begin{split}
        \mathcal{D}[\bm{u}(\bm{x});\bm{\kappa}] &= \bm{f}(\bm{x}), \; \bm{x} \in \Omega, \\
        \mathcal{B}_{k}[\bm{u}(\bm{x});\bm{\kappa}] &= \bm{g}_{k}(\bm{x}), \; \bm{x} \in \Gamma_{k} \subset \partial\Omega, \\
    \end{split} 
\end{equation}
where $\mathcal{D}$ is a differential operator acting on $\bm{u}(\bm{x})$ and parameterized by $\bm{\kappa}$, $\bm{u}(\bm{x}) \in \mathbb{R}^{d_{out}}$ is the hidden solution of the \ac{PDE} with $\bm{x}$ representing spatial points in the domain $\Omega \in \mathbb{R}^{d_{in}}$ and $\bm{f}(\bm{x})$ is a forcing term. The system of \acp{PDE} is further defined by a set of suitable boundary condition operators $\mathcal{B}_{k}$ acting on $\bm{u}(\bm{x})$ and a set of boundary functions $\bm{g}_{k}(\bm{x})$ which are imposed on the corresponding part $\Gamma_{k} \subset \partial\Omega$ on the boundary with $k=1, \dots, n_{bc}$. With \cref{eq:pinn_pde_bcs}, a general formulation of a variety of physical problems is provided.

When solving \acp{PDE} using \acp{PINN}, the \acp{PDE} are directly embedded into the loss function. Defining the loss function starts with approximating the hidden solution $\bm{u}(\bm{x})$ by an \ac{ANN} $\bm{\mathcal{N}}(\bm{x}; \bm{\theta})$, such that
\begin{equation}\label{eq:pinn_ann}
    \bm{u}(\bm{x}) \approx \bm{\mathcal{N}}(\bm{x}; \bm{\theta}).
\end{equation}
Additionally, in the inverse setting, the \ac{PDE} parameters $\bm{\kappa}$ become trainable parameters and are optimized during the \ac{PINN} training alongside the \ac{ANN} parameters $\bm{\theta}$ \cite{raissi_PhysicsinformedNeuralNetworks_2019}. Thus, the loss function introduced in \cref{subsec:artificial_neural_networks} depends on both $\bm{\theta}$ and $\bm{\kappa}$ as well as the training data $\bm{\mathcal{T}}$ and is defined as
\begin{equation}\label{eq:pinn_loss_function}
    \mathcal{L}(\bm{\theta},\bm{\kappa};\bm{\mathcal{T}}) := 
    \lambda_{pde}\mathcal{L}_{pde}(\bm{\theta},\bm{\kappa};\bm{\mathcal{T}}_{pde}) + \mathcal{L}_{bc}(\bm{\theta},\bm{\kappa};\bm{\mathcal{T}}_{bc}) +
    \lambda_{o}\mathcal{L}_{o}(\bm{\theta};\bm{\mathcal{T}}_{o}).
\end{equation}
The loss terms $\mathcal{L}_{pde}$, $\mathcal{L}_{bc}$ and $\mathcal{L}_{o}$ penalize the residual of the approximation $\bm{\mathcal{N}}(\bm{x};\bm{\theta})$ with respect to the \ac{PDE}, the boundary condition data and the observation data, respectively, and are defined as
\begin{equation}\label{eq:pinn_loss_terms}
    \begin{split}
		\mathcal{L}_{pde}(\bm{\theta},\bm{\kappa};\bm{\mathcal{T}}_{pde}) &= 
		\frac{1}{N_{pde}}\sum_{i=1}^{N_{pde}}\norm{\mathcal{D}[\bm{\mathcal{N}}(\bm{x}_{i};\bm{\theta});\bm{\kappa}]-\bm{f}(\bm{x}_{i})}^{2}_{2}, \\
		\mathcal{L}_{bc}(\bm{\theta},\bm{\kappa};\bm{\mathcal{T}}_{bc}) &= \lambda_{bc}^{k}\sum_{k}\frac{1}{N_{bc}^{k}}\sum_{j=1}^{N_{bc}^{k}}\norm{\mathcal{B}_{k}[\bm{\mathcal{N}}(\bm{x}_{j}^{k};\bm{\theta});\bm{\kappa}] - \hat{\bm{u}}_{j}^{k}}^{2}_{2}, \\
		\mathcal{L}_{o}(\bm{\theta};\bm{\mathcal{T}}_{o}) &= \frac{1}{N_{o}}\sum_{l=1}^{N_{o}}\norm{\bm{\mathcal{N}}(\bm{x}_{l};\bm{\theta}) - \hat{\bm{u}}_{l}}^{2}_{2}.
    \end{split}
\end{equation}
Here, the training data $\bm{\mathcal{T}}$ is comprised of three sets, $\bm{\mathcal{T}}_{pde}$, $\bm{\mathcal{T}}_{bc}$ and $\bm{\mathcal{T}}_{o}$. We refer to $\bm{\mathcal{T}}_{pde}$ as the $N_{pde}$ collocation points $\{\bm{x}_{i}\}_{i=1}^{N_{pde}}$ sampled from the domain $\Omega$. Furthermore, $\bm{\mathcal{T}}_{bc}$ is comprised of $k$ subsets, each with $N_{bc}^{k}$ training points $\{\bm{x}_{j}^{k},\hat{\bm{u}}_{j}^{k}\}_{j=1}^{N_{bc}^{k}}$ with $\hat{\bm{u}}_{j}^{k}=g_{k}(\bm{x}_{j}^{k})$ sampled from the boundary $\Gamma_{k}$ associated with the boundary condition. $\bm{\mathcal{T}}_{o}$ is set of $N_{o}$ data points $\{\bm{x}_{l}, \hat{\bm{u}}_{l}\}_{l=1}^{N_{o}}$ located in the domain $\Omega$ for which observations of the hidden function $\hat{\bm{u}}_{l}=\bm{u}(\bm{x}_{l})$ are available. The different loss terms $\mathcal{L}_{pde}$, $\mathcal{L}_{bc}$ and $\mathcal{L}_{o}$ can additionally be weighted by $\lambda_{pde}$, $\lambda_{bc}^{k}$ and $\lambda_{o}$, where $\lambda_{bc}^{k}$ can in principle be different for each $k$. For calculating $\mathcal{L}_{pde}(\bm{\theta}, \bm{\kappa};\bm{\mathcal{T}}_{pde})$ in \cref{eq:pinn_loss_terms}, the partial derivatives of the \ac{ANN} outputs with respect to the inputs are calculated using automatic differentiation \cite{baydin_AutomaticDifferentiationMachine_2018}. 

For inverse problems, the optimization problem is to find the optimal \ac{PINN} parameters $\bm{\theta^{*}}$ and material parameters $\bm{\kappa}^{*}$:
\begin{equation}\label{eq:pinn_optimization_problem_inverse}
    [\bm{\theta^{*}}, \bm{\kappa}^{*}]=\arg\min_{\bm{\theta},\bm{\kappa}}\mathcal{L}(\bm{\theta,\bm{\kappa}};\bm{\mathcal{T}}).
\end{equation}
Solving the problem defined in \cref{eq:pinn_optimization_problem_inverse} is called training. From \cref{eq:pinn_optimization_problem_inverse}, it can be seen that the \ac{PDE} parameters $\bm{\kappa}$ are estimated while simultaneously the forward solution is approximated by the \ac{ANN}. Thus, for inverse problems, \acp{PINN} follow the all-at-once approach. All-at-once approaches are outlined, e.g., in \cite{haber_PreconditionedAllAtOnce_2001}, \cite{schlintl_AllAtOnceMeetsBayesian_2021} and in \cite{guth_EnsembleKalmanFilter_2020}.


\section{Further development of \acp{PINN} towards a realistic data regime}\label{sec:method_development}
It is shown that standard \acp{PINN}, as presented in \cref{subsec:physics_informed_neural_networks}, fail in identifying material parameters from displacement data in a realistic regime without further modifications. Previous work, including \cite{haghighat_PhysicsinformedDeepLearning_2021} and \cite{zhang_PhysicsInformedNeuralNetworks_2020}, among others, have already proven \acp{PINN} to be generally capable of identifying material parameters from full-field displacement data. However, for this purpose, simplified assumptions were made which often do not match real-world conditions. These assumptions concern the domain size, the availability of stress data, the boundary conditions and the magnitude of both displacements and material parameters as well as the amount of noise in the data. In the following, we elaborate the necessary developments of the \ac{PINN} approach, which essentially aim at conditioning the optimization problem and presenting it in an appropriate formulation with respect to the goal of material model calibration. Some of the extensions described below were first introduced in a similar form in \cite{anton_IdentificationMaterialParameters_2022}. In this paper, we present further improvements and demonstrate the applicability in the two-dimensional regime. We expect that an extension to the three-dimensional regime is straightforward. Three-dimensional displacement data can be measured, e.g., by digital volume correlation \cite{buljac_DigitalVolumeCorrelation_2018}.

\subsection{Normalization of inputs and outputs}\label{subsec:method_normalization_pinn}
For \acp{PINN} to be able to approximate the displacement field from given experimental displacement data, both the inputs and outputs of the \ac{PINN} must be normalized. It is well known that the convergence of \ac{ANN} training can be accelerated by normalizing its inputs, where the mean values of all input features should each be close to zero according to \cite{y.a.lecun_EfficientBackProp_2012}. In addition, we found that in our use case, the output of the \ac{ANN} also needs to be normalized. Therefore, we approximate the hidden solution in \cref{eq:pde_balance_linear_momentum} by the normalized \ac{ANN} $\bar{\bm{\mathcal{N}}}$, defined as
\begin{equation}\label{eq:method_normalized_ann}
    \bar{\bm{\mathcal{N}}}(\bm{x}) := \bm{T}^{out}(\bm{\mathcal{N}}(\bm{T}^{in}(\bm{x}); \bm{\theta})).
\end{equation}
Since we use the hyperbolic tangent as activation function in the hidden layers, the output of these neurons is always in the range $[-1, 1]$. Therefore, we also map the input into the range $[-1,1]$ before we forward it to the \ac{ANN}. We then retransform the \ac{ANN} outputs back to the real output data range and thus force the \ac{ANN} to predict the outputs in the range $[-1, 1]$. The elementwise linear transformation $\bm{T}^{in}$ and re-transformation $\bm{T}^{out}$ are defined as
\begin{equation}\label{eq:method_ann_transformations}
    \begin{split}
        T_{i}^{in}(x_{i}) &= 2 \left(\frac{x_{i}-x_{i}^{min}}{x_{i}^{max}-x_{i}^{min}}\right) - 1, \; \forall \, i \in [1, d_{in}], \\
        T_{j}^{out}(\hat{u}_{j}) &= \left(\hat{u}_{j}^{max}-\hat{u}_{j}^{min}\right) \frac{\hat{u}_{j} + 1}{2} + \hat{u}_{j}^{min}, \; \forall \, j \in [1, d_{out}],.
    \end{split}
\end{equation}
The transformation and re-transformation is based on $x^{min}_{i}$, $x^{max}_{i}$, $\hat{u}^{min}_{j}$ and $\hat{u}^{max}_{j}$ which are the minimum and maximum values of the \ac{PINN} inputs and outputs for dimension $i$, respectively, and can be taken from the training data. As a consequence, the \ac{PINN} learns the mapping between the normalized inputs and the normalized outputs instead of learning the mapping between the real inputs and the real outputs. Applying the chain and product rule, we obtain the first and second derivatives of the output of the normalized \ac{ANN} with respect to the inputs in index notation:
\begin{equation}\label{eq:method_derivatives_normalized_ann}
    \begin{split}
        \frac{d \bar{\mathcal{N}}_{i}}{d x_{j}} &= \frac{\partial T^{out}_{i}}{\partial \mathcal{N}_{i}} \frac{\partial \mathcal{N}_{i}}{\partial T^{in}_{j}} \frac{d T^{in}_{j}}{d x_{j}}, \\
        \frac{d^{2} \bar{\mathcal{N}}_{i}}{d x_{j} x_{k}} &= \frac{\partial T^{out}_{i}}{\partial \mathcal{N}_{i}} \frac{\partial^{2} \mathcal{N}_{i}}{\partial T^{in}_{j} T^{in}_{k}} \frac{d T^{in}_{j}}{d x_{j}} \frac{d T^{in}_{k}}{d x_{k}}.
    \end{split}
\end{equation}
It is important to emphasize that at any time the real inputs are fed into the normalized \ac{ANN}. Likewise, the normalized \ac{ANN} always outputs the real outputs. Thus, the physics are not violated by the normalization when the outputs of the \ac{PINN} are derived according to the inputs during training.

\subsection{Conditioning of loss function}\label{subsec:method_conditioning_loss_function}
We also found that the loss function is ill-conditioned for small displacements, which prevents the \ac{PINN} to accurately approximate the displacement field. In real-world problems, the measured displacements may be in the order of magnitude of $10^{-3} \, \text{mm}$ or less. With displacements of this magnitude, even a relative error $RE = 100 \%$ of the approximation by the \ac{PINN} results in a data loss in the order of magnitude of $10^{-6}$ according to \cref{eq:pinn_loss_terms}. The small losses result in small gradients of the loss with respect to the \ac{ANN} parameters to be optimized. This in turn makes gradient based optimization difficult. Therefore, we first introduce the relative mean squared error. Applying this error metric in the data loss term in \cref{eq:pinn_loss_terms}, we obtain
\begin{equation}\label{eq:method_relative_mean_squred_error}
    \mathcal{L}_{o}(\bm{\theta};\bm{\mathcal{T}}_{o}) = \frac{1}{N_{o}}\sum_{l=1}^{N_{o}}\norm{(\bar{\bm{\mathcal{N}}}(\bm{x}_{l};\bm{\theta}) - \hat{\bm{u}}_{l}) \oslash \bm{u}^{char}}^{2}_{2},
\end{equation}
where $\bm{u}^{char}$ is a vector of characteristic displacements and $\oslash$ the Hadamard division operator. We propose to choose the characteristic displacements as the mean displacements from the training data set:
\begin{equation}\label{eq:method_characteristic_displacements}
    \bm{u}^{char} = \frac{1}{N_{o}} \sum_{i}^{N_{o}} \hat{\bm{u}}_{i}.
\end{equation}
The displacement may have different orders of magnitudes in the different dimensions. As a result, the contributions of the different dimensions to the data loss would vary depending on their magnitude. The relative mean squared error takes this into account and adaptively equalizes the contributions to the same level. Second, in order to increase the impact of the total data loss term $\mathcal{L}_{o}$ and to balance the loss terms, it is necessary to increase the weight $\lambda_{o}$ in \cref{eq:pinn_loss_function}. While the relative mean squared error used in \cref{eq:method_relative_mean_squred_error} is adaptive, the loss term weight is manually selected. Care must be taken to ensure that, on the one hand, the data loss term is given sufficient priority, but on the other hand, overfitting to the data is prevented. In principle, there are also approaches to adaptively weight the loss terms, such as \cite{wang_UnderstandingMitigatingGradient_2021,wang_WhenWhyPINNs_2022}. We have found, however, that adaptive loss term weighting may not improve the training dynamics for problems with many different loss terms, such as those encountered in solid mechanics. As can be seen in \cref{sec:1D_data} and \cref{sec:2D_data}, we also succeed in manually balancing the loss terms. A proper conditioning of the data loss term in combination with normalization allows \acp{PINN} to approximate even displacement fields of small magnitude. Note, however, that an accurate approximation of the displacement field is a necessary, but not a sufficient, criterion for identifying material parameters from realistic displacement data.

\subsection{Scaling of the parameters to be optimized}\label{subsec:method_scaling_parameters}
Due to different scales of the material and \ac{ANN} parameters, the calibration with the standard \ac{PINN} is not feasible for the optimizer. We examined the range of the optimized \ac{ANN} parameters $\bm{\theta}$ for the standard \ac{PINN} and found that for the analytical example in \cref{sec:1D_data} these are in the range $[-1457.0, 1881.38]$ , also see \cite{anton_IdentificationMaterialParameters_2022}. In contrast, the material parameters to be identified for construction steel are approximately $\nu = 0.3$ and $E = 210{,}000 \, \frac{\text{N}}{\text{mm}^{\text{2}}}$ or $K \approx 175{,}000.00 \, \frac{\text{N}}{\text{mm}^{\text{2}}}$ and $G \approx 80{,}769.23 \, \frac{\text{N}}{\text{mm}^{\text{2}}}$. Since both the \ac{ANN} and the material parameters are optimized simultaneously, this leads to a poorly scaled optimization problem \cite{nocedal_NumericalOptimization_2006}. We also assume that the optimizer encounters many local minima, which complicates the optimization. In order to obtain a similar scale for all parameters to be optimized, the real material parameters in the constitutive model will be replaced by
\begin{equation}\label{eq:method_scaled_parameter}
    \begin{split}
        \kappa = (1.0 + \alpha_{\kappa}) \kappa_{est}.
    \end{split}
\end{equation}
Here $\alpha_{\kappa}$ and $\kappa_{est}$ are a correction factor and an initial estimate for the material parameter to be identified, respectively. From now on, the correction factor is identified instead of the material parameter. If we provide the exact Young's modulus, then the optimized parameters of the enhanced \ac{PINN} $\bm{\theta}$ are now in the range $[-5.34, 4.40]$ and the correction factor $\alpha_{\kappa}$ very close to $1$, as we will see in \cref{sec:1D_data}. Thus, the parameters to be optimized have a similar scale and the optimization problem is well posed. Unless otherwise specified, the correction factors are initialized with zeros throughout this paper.

\subsection{Problem reformulation}\label{subsec:method_problem_reformulation}
The accuracy as well as the robustness of the identified material parameters strongly depend on how the optimization problem is formulated. The advantage of Young's modulus and Poisson's ratio over alternative elasticity parameters is that they are often easier to interpret. However, in attempting to identify them directly, we observed that the results are less accurate and, more importantly, less robust than when the constitutive model is formulated with respect to bulk and shear modulus. We made this observation for the case where we provide an estimate of $\nu_{est} = 0.45$ for Poisson's ratio. For some random initializations of the \ac{PINN} parameters, this estimate results in a failure of the calibration. Therefore, we reformulated the optimization problem in terms of the material parameters to be identified. We first converted the estimated Young's modulus and Poisson's ratio into the corresponding estimates for bulk and shear modulus according to \cref{eq:bulk_shear_modulus}. With the scaling of the parameters proposed in \cref{subsec:method_scaling_parameters}, the modified material parameters in \cref{eq:pde_balance_linear_momentum} then become $\bar{\bm{\kappa}}=[\alpha_{K}, \alpha_{G}]$. In one-dimensional problems, the modified material parameter remains $\bar{\kappa} = \alpha_{E}$. Finally, the identified bulk and shear modulus are converted back to Young's modulus and Poisson's ratio. We assume that this formulation of the optimization problem is better posed because it introduces a constraint on the Poisson's ratio. As soon as the Poisson's ratio approaches the incompressibility limit of $\nu=0.5$, the bulk modulus would approach infinity. Critical values $\nu \geq 0.5$ can thus be prevented and the optimization becomes more robust as a result.

\subsection{Incorporating balance of internal and external work}\label{subsec:method_balance_work}
In order to be able to identify material parameters, we need to account either for the global or at least some local force information. As soon as no volume force is present, at least some information on the force boundary conditions must be taken into account so that the material parameter is uniquely identifiable. Otherwise, the inverse problem would be ill-posed. There would exist an infinite number of combinations of material parameters and force situations, which lead to the same displacement field. In the standard \ac{PINN} formulation, the accuracy of the material parameters depends strongly on the accuracy of the displacement field approximation in regions where the stress boundary conditions are applied. A large error in the approximation near the boundaries then inevitably leads to a large error in the identified material parameters. This can cause problems especially when dealing with erroneous or noisy data. In order to remove this local dependency, instead of the stress boundary conditions, we consider the internal and external energy of the mechanical system. Contrary to \cite{samaniego_EnergyApproachSolution_2020}, however, we do not minimize the total potential energy of the system but use the balance of internal and external work equivalent to the principle of conservation of mechanical energy. According to \cref{eq:balance_work}, we enforce the balance of internal and external work by the loss term
\begin{equation}\label{eq:method_loss_term_work}
    \mathcal{L}_{W}(\bm{\theta},\bar{\bm{\kappa}};\bm{\mathcal{T}}_{W}) = \\
	W_{int}(\bm{\theta}, \bar{\bm{\kappa}}; \bm{\mathcal{T}}_{W_{int}})
	- W_{ext}(\bm{\theta}; \bm{\mathcal{T}}_{W_{ext}}),
\end{equation}
where the training data $\bm{\mathcal{T}}_{W}$ is comprised of $\bm{\mathcal{T}}_{W_{int}}$ and $\bm{\mathcal{T}}_{W_{ext}}$. For the approximation of the domain and boundary integrals in the calculation of the internal and external work, we adopt the Monte-Carlo integration. The internal work can thus be calculated approximately by
\begin{equation}\label{eq:method_internal_work}
    W_{int}(\bm{\theta}, \bar{\bm{\kappa}}; \bm{\mathcal{T}}_{W_{int}}) \approx \\
    \frac{1}{2} \frac{V_{\Omega}}{N_{W_{int}}} \sum_{m=1}^{N_{W_{int}}}  \bm{\sigma}(\bar{\bm{\mathcal{N}}}(\bm{x}_{m};\bm{\theta}) ;\bm{\kappa}) : \bm{\varepsilon}(\bar{\bm{\mathcal{N}}}(\bm{x}_{m};\bm{\theta})),
\end{equation}
where $\bm{\mathcal{T}}_{W_{int}}$ is a set of $N_{W_{int}}$ collocation points $\{\bm{x}_{m}\}_{m=1}^{N_{W_{int}}}$ sampled from the domain $\Omega$. In one and two dimensions, the volume $V_{\Omega}$ corresponds to the length and the area of the domain $\Omega$, respectively. The first derivative of the \ac{PINN} output with respect to its inputs, which is needed to calculate the strains $\bm{\varepsilon}(\bm{u(\bm{x})})$ from \cref{eq:kinematic_law}, can again be calculated using automatic differentiation. In the absence of body forces, the external work can be approximated by
\begin{equation}\label{eq:method_external_work}
    W_{ext}(\bm{\theta}; \bm{\mathcal{T}}_{W_{ext}}) \approx \\
    \frac{1}{2} \frac{V_{\partial\Omega}}{N_{W_{ext}}} \sum_{n=1}^{N_{W_{ext}}} \bm{t}(\bm{x_{n}}) \cdot \bar{\bm{\mathcal{N}}}(\bm{x_{n}};\bm{\theta})
\end{equation}
Here, $\bm{\mathcal{T}}_{W_{ext}}$ is a set of $N_{W_{ext}}$ collocation points $\{\bm{x}_{n}\}_{n=1}^{N_{W_{ext}}}$ sampled from the boundary $\partial\Omega$. We assume that the traction $\bm{t}(\bm{x})$ is known at least for the collocation points in $\bm{\mathcal{T}}_{W_{ext}}$ and that the collocation points are distributed as evenly as possible in the domain. Thus, we shift the dependence of the material parameter identification from the local accuracy of the displacement approximation in the boundary regions to the accuracy of the whole domain.

\subsection{Summary of the method developments}\label{subsec:method_summary}
Based on the general \ac{PINN} formulation reviewed in \cref{subsec:physics_informed_neural_networks} and the extensions introduced above, the \ac{PINN} and the associated optimization problem for calibrating the linear-elastic material model are defined as follows: In two dimensions, the displacement field is approximated by two independent \acp{ANN}, such that
\begin{equation}\label{eq:method_summary_pinn_2D}
    \bm{u}(\bm{x}) 
    \approx \bar{\bm{\mathcal{N}}}(\bm{x};\bar{\bm{\theta}}) \approx
        \begin{bmatrix}
            \bar{\mathcal{N}}_{x}(\bm{x}; \bm{\theta}_{x}) \\
            \bar{\mathcal{N}}_{y}(\bm{x}; \bm{\theta}_{y})
        \end{bmatrix},
\end{equation}
with $\bar{\bm{\theta}}=[\bm{\theta}_{x}, \bm{\theta}_{y}]$. According to \cite{haghighat_PhysicsinformedDeepLearning_2021}, this leads to a more accurate approximation then using one \ac{ANN} with two outputs, since the cross-dependency between the outputs may not be accurately represented by a single network. This cross-dependency results from the kinematic law in \cref{eq:kinematic_law} and the constitutive model. The loss function of the inverse problem is given by
\begin{equation}\label{eq:method_summary_loss_function_2D}
    \begin{split}
        \mathcal{L}(\bar{\bm{\theta}},\bar{\bm{\kappa}};\bm{\mathcal{T}}) &:=
        \mathcal{L}_{pde}(\bar{\bm{\theta}},\bar{\bm{\kappa}};\bm{\mathcal{T}}_{pde})
        + \mathcal{L}_{W}(\bar{\bm{\theta}},\bar{\bm{\kappa}};\bm{\mathcal{T}}_{W}) \\
        &\quad\,\,+ \lambda_{o}\mathcal{L}_{o_{x}}(\bm{\theta}_{x};\bm{\mathcal{T}}_{o}) 
        + \lambda_{o}\mathcal{L}_{o_{y}}(\bm{\theta}_{y};\bm{\mathcal{T}}_{o}),
    \end{split}
\end{equation}
with the loss terms defined as
\begin{equation}\label{eq:method_summary_loss_terms_2D}
    \begin{split}
		\mathcal{L}_{pde}(\bar{\bm{\theta}},\bar{\bm{\kappa}};\bm{\mathcal{T}}_{pde}) &=
		\frac{1}{N_{pde}}\sum_{i=1}^{N_{pde}}\norm{\operatorname{div}\bm{\sigma}(\bar{\bm{\mathcal{N}}}(\bm{x}_{i};\bar{\bm{\theta}}); \bar{\bm{\kappa}})}^{2}_{2}, \\
        \mathcal{L}_{W}(\bar{\bm{\theta}},\bar{\bm{\kappa}};\bm{\mathcal{T}}_{W}) &=
	    \frac{V_{\Omega}}{N_{W_{int}}} \sum_{m=1}^{N_{W_{int}}}  \bm{\sigma}(\bar{\bm{\mathcal{N}}}(\bm{x}_{m};\bar{\bm{\theta}}) ;\bar{\bm{\kappa}}) : \bm{\varepsilon}(\bar{\bm{\mathcal{N}}}(\bm{x}_{m};\bar{\bm{\theta}})) \\
	    &\quad\, - \frac{V_{\partial\Omega}}{N_{W_{ext}}} \sum_{n=1}^{N_{W_{ext}}} \bm{t}(\bm{x_{n}}) \cdot \bar{\bm{\mathcal{N}}}(\bm{x_{n}};\bar{\bm{\theta}}), \\
		\mathcal{L}_{o_{x}}(\bm{\theta}_{x};\bm{\mathcal{T}}_{o}) &= \frac{1}{N_{o}}\sum_{l=1}^{N_{o}}\norm{\frac{\mathcal{N}_{x}(\bm{x}_{l};\bm{\theta}_{x}) - \hat{u}_{x}^{l}}{u_{x}^{char}}}^{2}_{2}, \\
		\mathcal{L}_{o_{y}}(\bm{\theta}_{y};\bm{\mathcal{T}}_{o}) &= \frac{1}{N_{o}}\sum_{l=1}^{N_{o}}\norm{\frac{\mathcal{N}_{y}(\bm{x}_{l};\bm{\theta}_{y}) - \hat{u}_{y}^{l}}{u_{y}^{char}}}^{2}_{2},
    \end{split}
\end{equation}
where $[\hat{u}_{x}^{l}, \hat{u}_{y}^{l}]^{T} = \bm{u}(x_{l})$ is known from the observation data.

In one dimension, the displacement is approximated by only one \ac{ANN}, such that
\begin{equation}\label{eq:method_summary_pinn_1D}
    u(x) \approx \bar{\mathcal{N}}(x;\bm{\theta}), 
\end{equation}
and the loss function simplifies to
\begin{equation}\label{eq:method_summary_loss_function_1D}
    \mathcal{L}(\bm{\theta},\bar{\kappa};\bm{\mathcal{T}}) :=
    \mathcal{L}_{pde}(\bm{\theta},\bar{\kappa};\bm{\mathcal{T}}_{pde})
    + \mathcal{L}_{W}(\bm{\theta},\bar{\kappa};\bm{\mathcal{T}}_{W})
    + \lambda_{o}\mathcal{L}_{o}(\bm{\theta};\bm{\mathcal{T}}_{o}) 
\end{equation}
Correspondingly, the loss terms are then defined as
\begin{equation}\label{eq:method_summary_loss_terms_1D}
    \begin{split}
		\mathcal{L}_{pde}(\bm{\theta},\bar{\kappa};\bm{\mathcal{T}}_{pde}) &=
		\frac{1}{N_{pde}}\sum_{i=1}^{N_{pde}}\left(E\frac{\partial}{\partial^{2}x}\bar{\mathcal{N}}(x_{i};\bm{\theta})\right)^{2}, \\
		\mathcal{L}_{W}(\bm{\theta},\bar{\kappa};\bm{\mathcal{T}}_{W}) &=
	    \frac{V_{\Omega}}{N_{W_{int}}} \sum_{m=1}^{N_{W_{int}}}  \sigma(\bar{\mathcal{N}}(x_{m};\bm{\theta}) ;\bar{\kappa}) \cdot \varepsilon(\bar{\mathcal{N}}(x_{m};\bm{\theta}))\\
	    &\quad\, - V_{\partial\Omega} \cdot t(x_{n}) \cdot \bar{\mathcal{N}}(x_{n};\bm{\theta}), \\
		\mathcal{L}_{o}(\bm{\theta};\bm{\mathcal{T}}_{o}) &= \frac{1}{N_{o}}\sum_{l=1}^{N_{o}}\left(\frac{\mathcal{N}(\bm{x}_{l};\bm{\theta})- \hat{u}^{l}}{u^{char}}\right)^{2}. \\
    \end{split}
\end{equation}
Here, $x_{n}$ is the coordinate of the boundary where traction is applied. 

Unless otherwise specified, the weight for the data loss terms $\lambda_{o}$ and the characteristic displacements $\bm{u}^{char}$ are set to $\lambda_{o} = 10^{5}$ and according to \cref{eq:method_characteristic_displacements}. In addition, we use the same number and location of collocation points $N_{col}$ to determine the \ac{PDE} residual $\lambda_{pde}$ and to determine the internal work $W_{int}$, such that $N_{pde} = N_{W_{int}} = N_{col}$.


\section{Parameter identification from one-dimensional displacement data}\label{sec:1D_data}
We first demonstrate the enhanced method for one-dimensional displacement data. Using analytical data, the method is validated for a realistic data regime and a sensitivity analysis is conducted with respect to the initial estimate of the material parameter to be identified. We then show that \acp{PINN} are, in principle, also able to identify the Young's modulus from one-dimensional experimental displacement data.

\subsection{Analytical displacement data}\label{subsec:1D_analytical}
\begin{figure}[htb]
    \centering
    \begin{tikzpicture}
        \newcommand\XO{0}
        \newcommand\YO{0}
        \draw (\XO,\YO) -- (\XO,\YO+4) -- (\XO+0.5,\YO+4) -- (\XO+0.5,\YO) -- (\XO,\YO);
        \draw[->, thick] (\XO,\YO) -- ++(270:0.5);
        \draw[->, thick] (\XO+0.25,\YO) -- ++(270:0.5);
        \draw[->, thick] (\XO+0.5,\YO) -- ++(270:0.5);
        \draw[thick] (\XO,\YO-0.5) -- (\XO+0.5,\YO-0.5);
        \draw[thick,->] (\XO+1.5, \YO+4) -- (\XO+1.5, \YO+3) node[anchor=south west] {x};
        \draw[<->] (\XO+0.75,\YO+4) -- (\XO+0.75,\YO+0) node[midway, below, rotate=90] {$L = 100 \, \text{mm}$};
        \node[left] at (\XO-0.125,\YO+4) {$\bar{u} = 0$};
        \node[left] at (\XO-0.25,\YO-0.25) {$\bar{t}=100 \, \frac{\text{N}}{\text{mm}^{\text{2}}}$};
    \end{tikzpicture} 
    \caption{Geometry and boundary conditions of a stretched rod under tension load. The stretched rod is clamped at the upper end and a traction is applied at the lower end. Body forces are neglected.}
    \label{fig:setup_1D_analytical}
\end{figure}
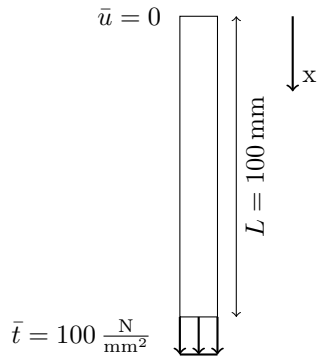

As a first test case, we consider analytical displacement data for a stretched rod. The stretched rod has a length of $L = 100 \, \text{mm}$ and its upper end is clamped. A traction of $\bar{t} = 100 \, \frac{\text{N}}{\text{mm}^{\text{2}}}$ is applied at the free end, causing a deformation. External body forces, such as gravity, are neglected. We assume linear-elastic material with Young's modulus $E_{true} = 210{,}000 \, \frac{\text{N}}{\text{mm}^{\text{2}}}$. From the analytical solution, we calculate the displacements at $N_{o} = 128$ equidistantly sampled points to train the \ac{PINN} and at another $N_{val} = 1{,}024$ points to validate it. The collocation points for calculating the \ac{PDE} residual as well as the internal work during training have the same number and location as the data points, such that $N_{col} = N_{o} = 128$. The geometry and the boundary conditions are schematically illustrated in \cref{fig:setup_1D_analytical}.

For the approximation of the displacement, we use a fully connected \ac{FFNN} as defined in \cref{subsec:artificial_neural_networks} and set the architecture and training hyperparameters as follows: The \ac{ANN} has $2$ hidden layers, each with $8$ neurons. The hyperbolic tangent is used as activation function in the hidden layers. For optimization of the \ac{ANN} parameters and the correction factor of Young's modulus we use the BFGS \cite{broyden_ConvergenceClassDoublerank_1970, fletcher_NewApproachVariable_1970, shanno_ConditioningQuasiNewtonMethods_1970, goldfarb_FamilyVariablemetricMethods_1970} optimizer. The weights are initialized using Glorot normal initialization \cite{glorot_UnderstandingDifficultyTraining_2010}. The biases and the correction factor introduced in \cref{sec:method_development} are initialized with zeros. Optimization is performed in full batch mode. Since the chosen hyperparameters have proven to be suitable for this use case, no hyperparameter optimization is conducted.

\begin{figure}[htb]
    \centering
    \begin{subfigure}{0.49\textwidth}
        \centering
        \includegraphics[width=0.95\linewidth]{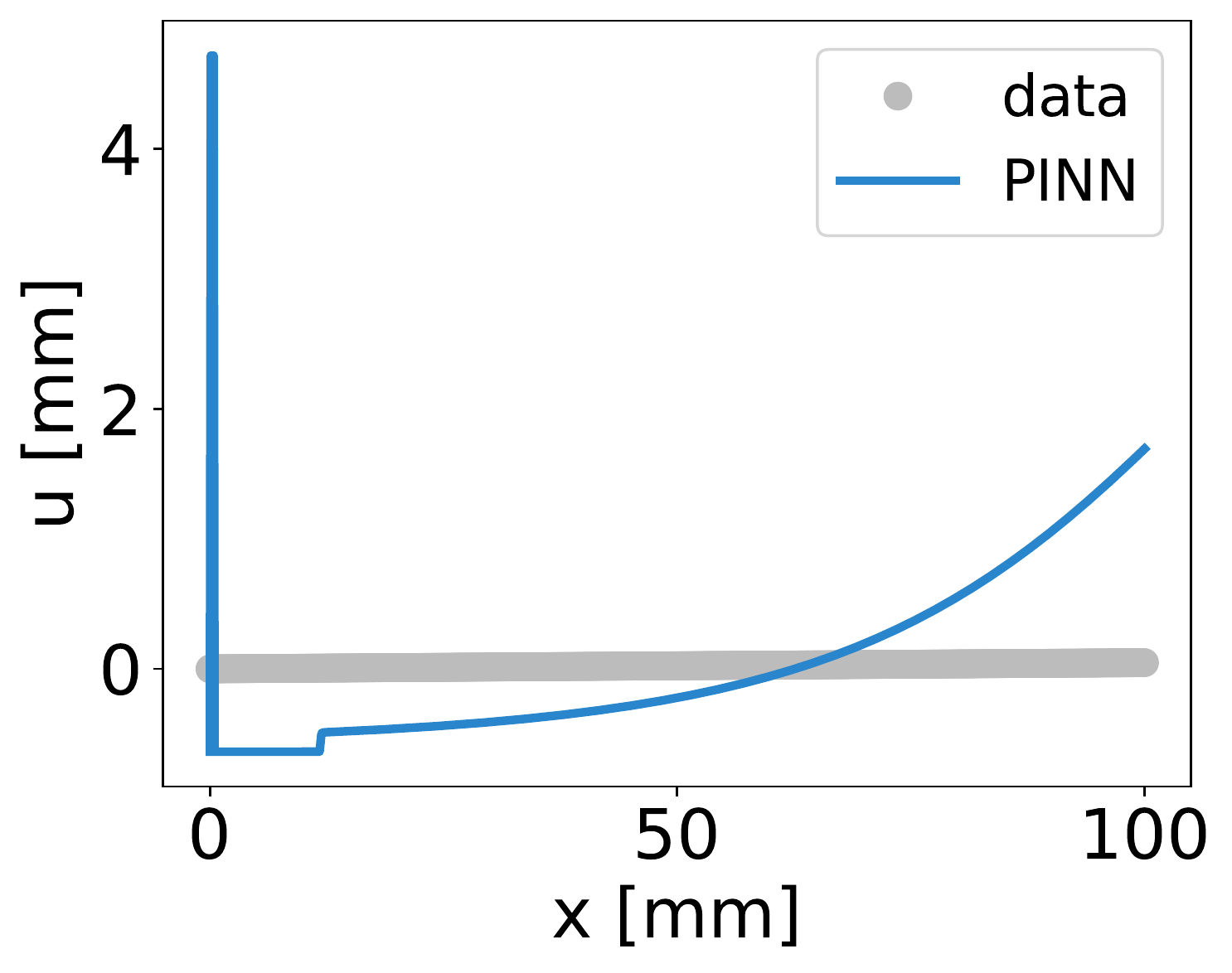}
        \caption{}
    \end{subfigure}
    \begin{subfigure}{0.49\textwidth}
        \centering
        \includegraphics[width=1\linewidth]{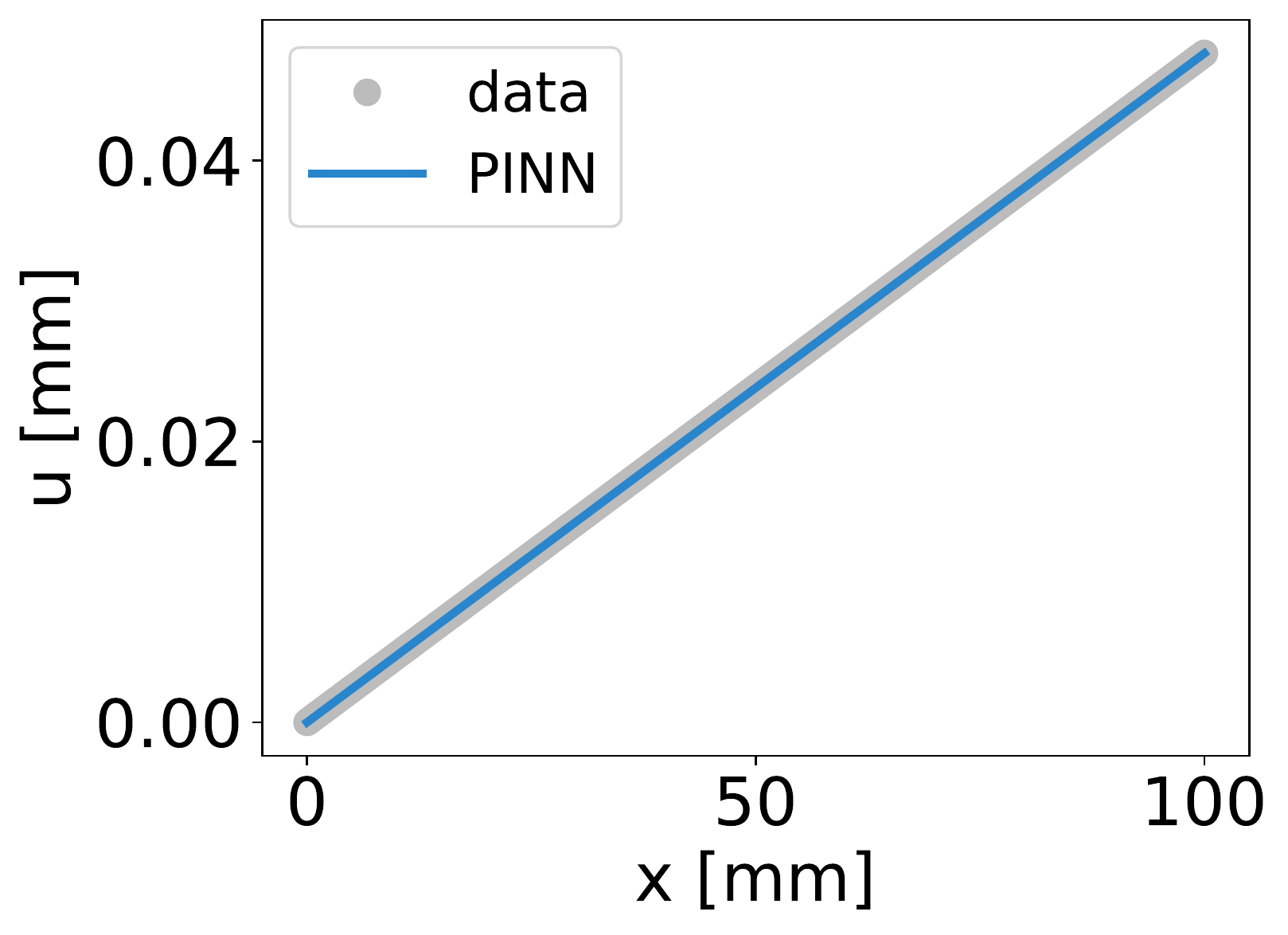}
        \caption{}
    \end{subfigure}
    \caption{\textbf{Standard \ac{PINN} vs. enhanced \ac{PINN}:} Displacement of the stretched rod approximated by a) a standard \ac{PINN} and b) the enhanced \ac{PINN}. The comparison demonstrates that the standard \ac{PINN} is not able to approximate the displacement of the stretched rod from the given data in a realistic regime.}
    \label{fig:1D_analytical_comparison_displacement}
\end{figure}

In order to demonstrate the effectiveness of the enhanced method, we try to identify the Young's modulus from the analytical displacement data with both the standard \ac{PINN} with Neumann boundary condition and the enhanced \ac{PINN}. For the enhanced \ac{PINN}, we compute the characteristic displacement $u^{char} = 0.0238 \, \text{mm}$ and set the data loss weight to $\lambda_{o} = 10^{5}$. The initial estimate is first set to the exact Young's modulus $E_{true} = 210{,}000 \, \frac{\text{N}}{\text{mm}^{\text{2}}}$ which we used for data generation. In \cref{fig:1D_analytical_comparison_displacement}, the displacement approximation for both the standard and the enhanced \acp{PINN} are shown. The comparison underlines that the standard \ac{PINN} is not capable to accurately approximate the displacement of the stretched rod. As a consequence, the identified Young's modulus is with $E_{ident}=1{,}442.35 \, \frac{\text{N}}{\text{mm}^{\text{2}}}$ far from the correct value. In contrast, the enhanced \ac{PINN} succeeds in identifying a Young's modulus of $E_{ident}=209{,}999.99 \, \frac{\text{N}}{\text{mm}^{\text{2}}}$ with a \acf{RE} of $RE_{E}=-4.45 \cdot 10^{-6} \%$. In addition, the relative $L^{2}\text{-norm}$ ($\text{r}L^{2}\text{-norm}$) of the displacement approximation is $\text{r}L^{2}=1.5577 \cdot 10^{-7}$.

\begin{figure}[htb]
    \centering
    \includegraphics[scale=0.50]{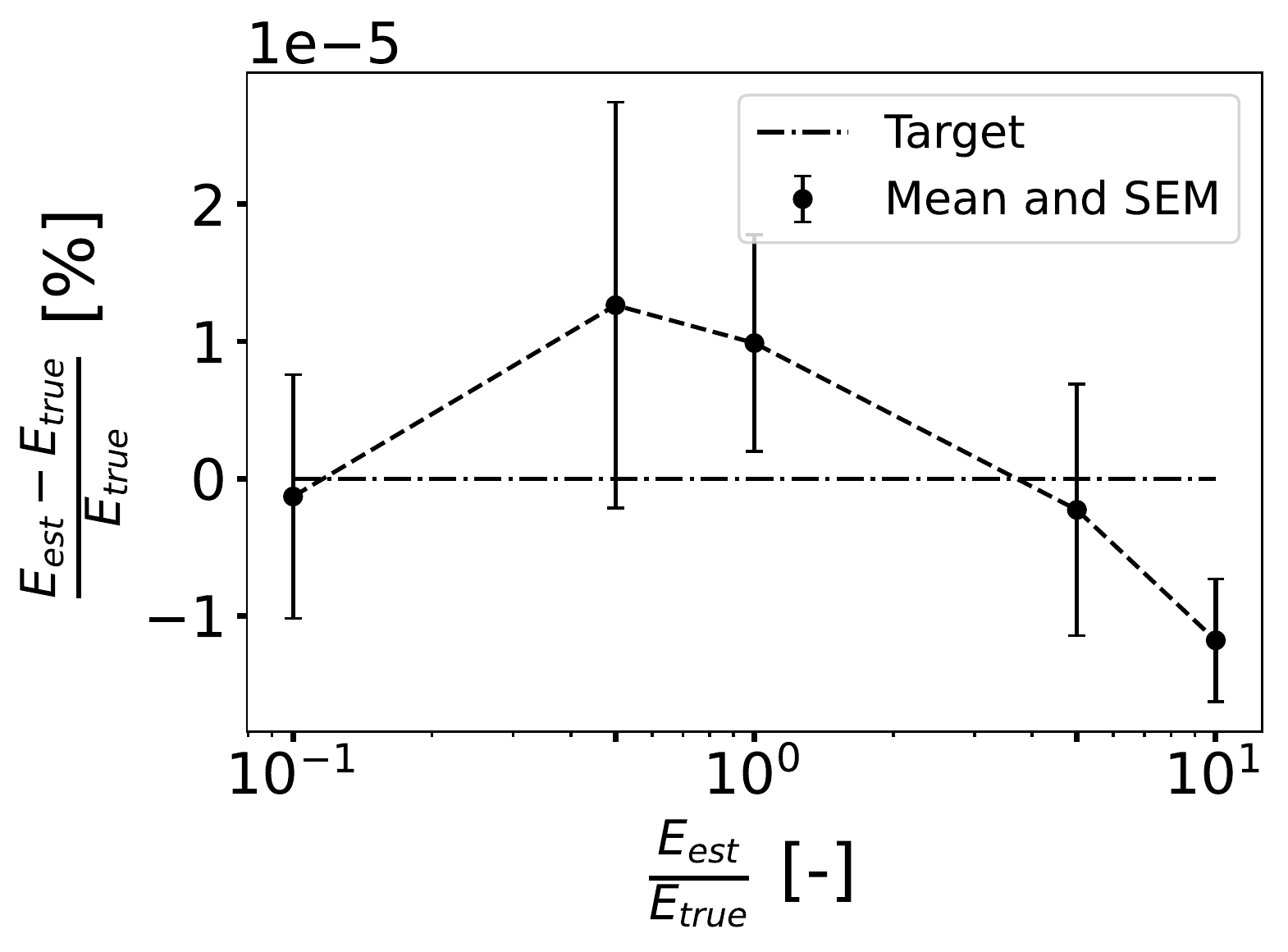}
    \caption{\textbf{Sensitivity analysis:} Error sensitivity of the identified Young's modulus $E$ with respect to the initial estimate. We considered initial estimates in the range $[10 \%, 1{,}000 \%]$ of the value used for data generation. The simulation for an estimated value was repeated a total of $10$ times with different randomly initialized \ac{ANN} parameters. The figure shows the mean values and \acfp{SEM} of the relative errors calculated from the individual runs.}
    \label{fig:1D_analytical_sensitivity}
\end{figure}

The results presented above, and the method in general, are only useful as long as the results are satisfactory even if the exact material parameters are not known. Hence, we investigate the sensitivity of the relative error of the identified Young's modulus with respect to the initial estimate. We perform a sensitivity analysis in which we solve the inverse problem for several initial estimates in the range $E_{est} = [10 \%, 1{,}000 \%] \cdot E_{true}$ of the Young's modulus used for data generation. The \ac{ANN} parameters are initialized identically for each simulation of one analysis. Since we found that the optimization result depends on the initialization of the \ac{PINN} parameters, we repeated the sensitivity analysis for a total of $10$ different random initializations. From the $10$ individual results for each initial estimate, we calculate the mean error and the \ac{SEM}. The latter is defined as
\begin{equation}\label{eq:SEM}
    SEM = \frac{\sigma}{\sqrt{n_{s}}}.
\end{equation}
Here, $\sigma$ is the standard deviation of the results and $n_{s}$ is the number of samples, where in our case $n_{s}=10$. The results in \cref{fig:1D_analytical_sensitivity} suggest that the error of the identified material parameter is not sensitive to the initial estimate within a reasonable range.

\subsection{Experimental displacement data}\label{subsec:1D_experimental}
We test the enhanced \ac{PINN} using one-dimensional experimental displacement data. For the material model calibration, we use the same \ac{PINN} architecture and hyperparameters as in \cref{subsec:1D_analytical} and set the data loss weight to $\lambda_{o}=10^{4}$. We choose the initial estimate as $E_{est}=210{,}000 \, \frac{\text{N}}{\text{mm}^{\text{2}}}$. The characteristic displacement is $u^{char} = 0.0579 \, \text{mm}$. As a reference solution, we use the prediction of a \ac{LS-FEM} simulation. For more information on this approach, we refer to \cite{hartmann_MaterialParameterIdentification_2021, hartmann_MaterialParameterIdentification_2021a}.

\begin{figure}[htb]
    \centering
    \begin{tikzpicture}[scale=0.95]
        \newcommand\XO{0}
        \newcommand\YO{0}
        \coordinate (A) at (\XO,\YO);
        \coordinate (B) at (\XO+2.5,\YO);
        \coordinate (C) at (\XO+3.25,\YO+0.25);
        \coordinate (D) at (\XO+7.75,\YO+0.25);
        \coordinate (E) at (\XO+8.5,\YO);
        \coordinate (F) at (\XO+11,\YO);
        \coordinate (G) at (\XO+11,\YO+1.5);
        \coordinate (H) at (\XO+8.5,\YO+1.5);
        \coordinate (I) at (\XO+7.75,\YO+1.25);
        \coordinate (J) at (\XO+3.25,\YO+1.25);
        \coordinate (K) at (\XO+2.5,\YO+1.5);
        \coordinate (L) at (\XO,\YO+1.5);
        \draw (A) -- (B);
        \draw (C) arc (90:126.87:1.25);
        \draw (C) -- (D);
        \draw (D) arc (90:53.13:1.25);
        \draw (E) -- (F);
        \draw (F) -- (G);
        \draw (G) -- (H);
        \draw (I) arc (270:306.87:1.25);
        \draw (I) -- (J);
        \draw (J) arc (270:233.13:1.25);
        \draw (K) -- (L);
        \draw (L) -- (A);
        \draw[<->] (\XO-0.25,\YO) -- (\XO-0.25,\YO+1.5) node[midway, below, rotate=270] {30 mm};
        \draw[<->] (\XO+11.25,\YO+0.25+0.01875) -- (\XO+11.25,\YO+1.25-0.01875) node[midway, above, rotate=270] {$W = 19.25 \, \text{mm}$};
        \draw[<->] (\XO,\YO+1.75) -- (\XO+2.5,\YO+1.75) node[midway, above] {50 mm};
        \draw[<->] (\XO+3.5,\YO-0.25) -- (\XO+7.5,\YO-0.25) node[midway, above] {$L= 80 \text{mm}$};
        \draw[<->] (\XO+3.25,\YO+1.75) -- (\XO+7.75,\YO+1.75) node[midway, above] {90 mm};
        \draw[<->] (\XO,\YO-0.75) -- (\XO+11,\YO-0.75) node[midway, above] {220 mm};
        \draw[<-] (\XO+3.25-0.395,\YO+0.25-0.064) -- ++(288.43:0.5) node[left] {R 25 mm};
        \draw[red] (\XO+3.5,\YO+0.25) -- (\XO+7.5,\YO+0.25) -- (\XO+7.5,\YO+1.25)  -- (\XO+3.5,\YO+1.25) --(\XO+3.5,\YO+0.25);
        \draw[red, densely dotted] (\XO+3.5,\YO+0.75) -- (\XO+7.5,\YO+0.75);
    \end{tikzpicture} 
    \caption{Geometry of the tensile specimen. The displacements were measured by a \ac{DIC} system for the area which is outlined in red. For the material model calibration, the two-dimensional displacement data was interpolated onto a total of $N_{o} = 161$ equidistant locations on the central axis of the specimen.}
    \label{fig:setup_1D_experimental}
\end{figure}
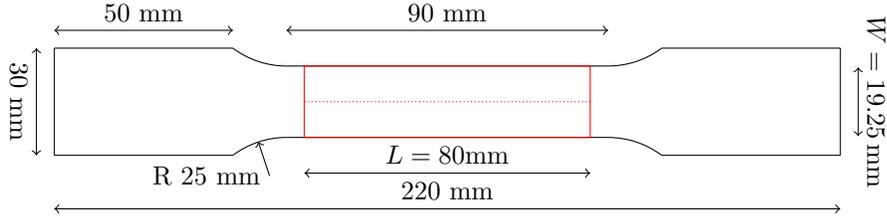

The experimental displacement data were measured in an uniaxial tensile test with the following setting: A specimen of TS275 steel was used, the geometry of which is shown in \cref{fig:setup_1D_experimental}. The test was performed displacement-controlled using the tensile testing machine Z100 from ZwickRoell GmbH \& Co. KG. While one end of the specimen was clamped, the testing machine pulled on the other end up to an averaged axial strain of $\varepsilon^{mean} = 4.69 \cdot 10^{-2} \%$. Thus, the strain is still in the assumed linear-elastic range of the material under consideration. In addition, the displacements were measured by the \ac{DIC} system Aramis 12M from Carl Zeiss GOM Metrology GmbH for an area of $L = 80 \, \text{mm}$ length and $W = 19.25 \, \text{mm}$ width in the center of the specimen. The displacements measured in two dimensions were then interpolated to a total of $N_{o} = 161$ equidistant locations on the central axis of the specimen, as illustrated in \cref{fig:setup_1D_experimental}. As in the analytical test case, the data points are also used as collocation points in the \ac{PDE} loss term and for calculating the internal work when training the \ac{PINN}, such that $N_{col} = N_{o} = 161$. Apart from the displacements, the force applied to initiate the displacements was measured by a force gauge. The force and the cross-sectional area of the specimen result in a traction $t = 212.55 \, \frac{\text{N}}{\text{mm}^{\text{2}}}$. The external work is then calculated from the reaction forces on the left and right boundaries of the free body of the considered area. The rigid body displacements are subtracted from the measured displacements.

\begin{figure}[htb]
    \centering
    \begin{subfigure}{0.49\textwidth}
        \centering
        \includegraphics[width=0.95\linewidth]{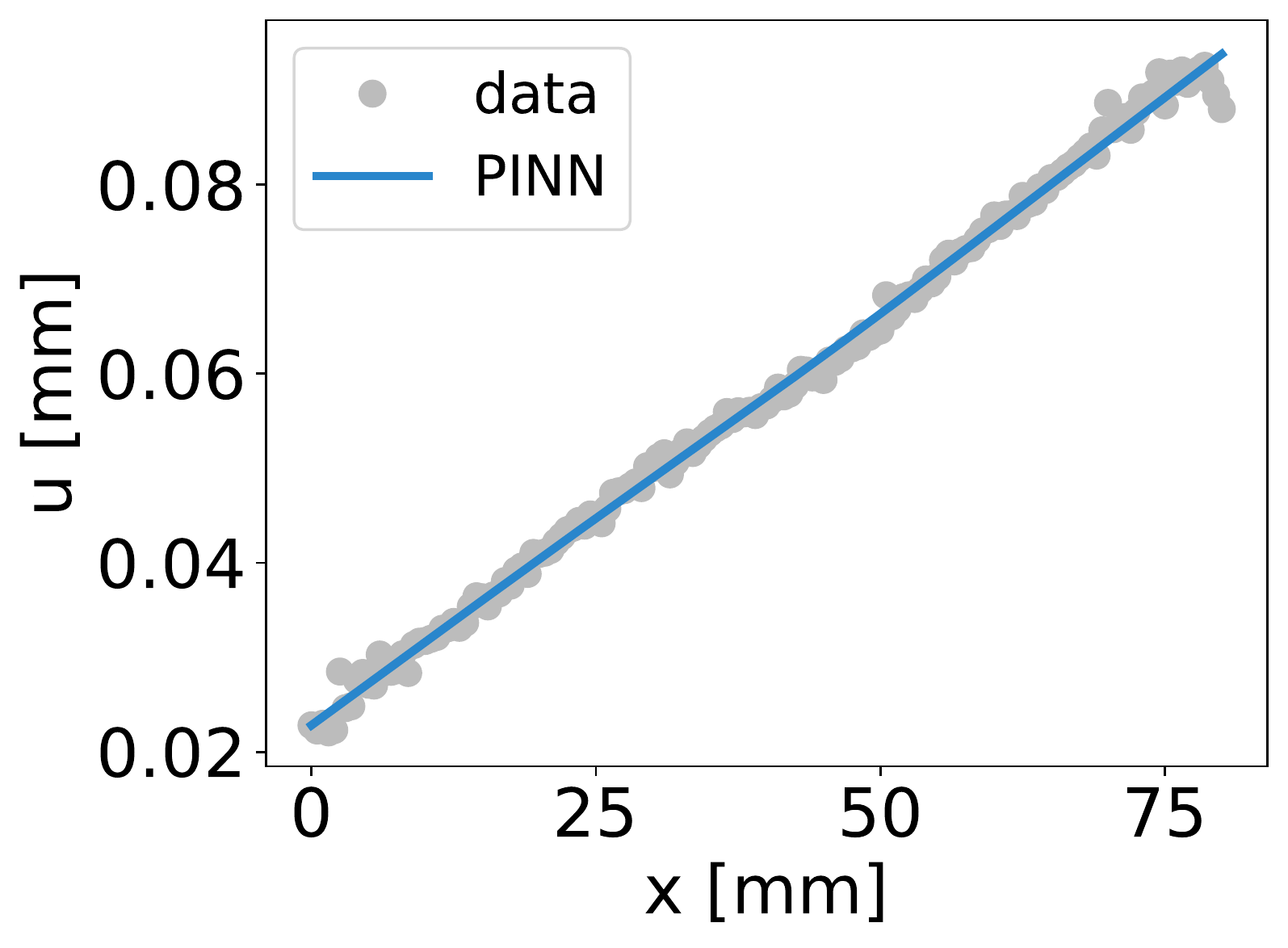}
        \caption{}
    \end{subfigure}
    \begin{subfigure}{0.49\textwidth}
        \centering
        \includegraphics[width=0.95\linewidth]{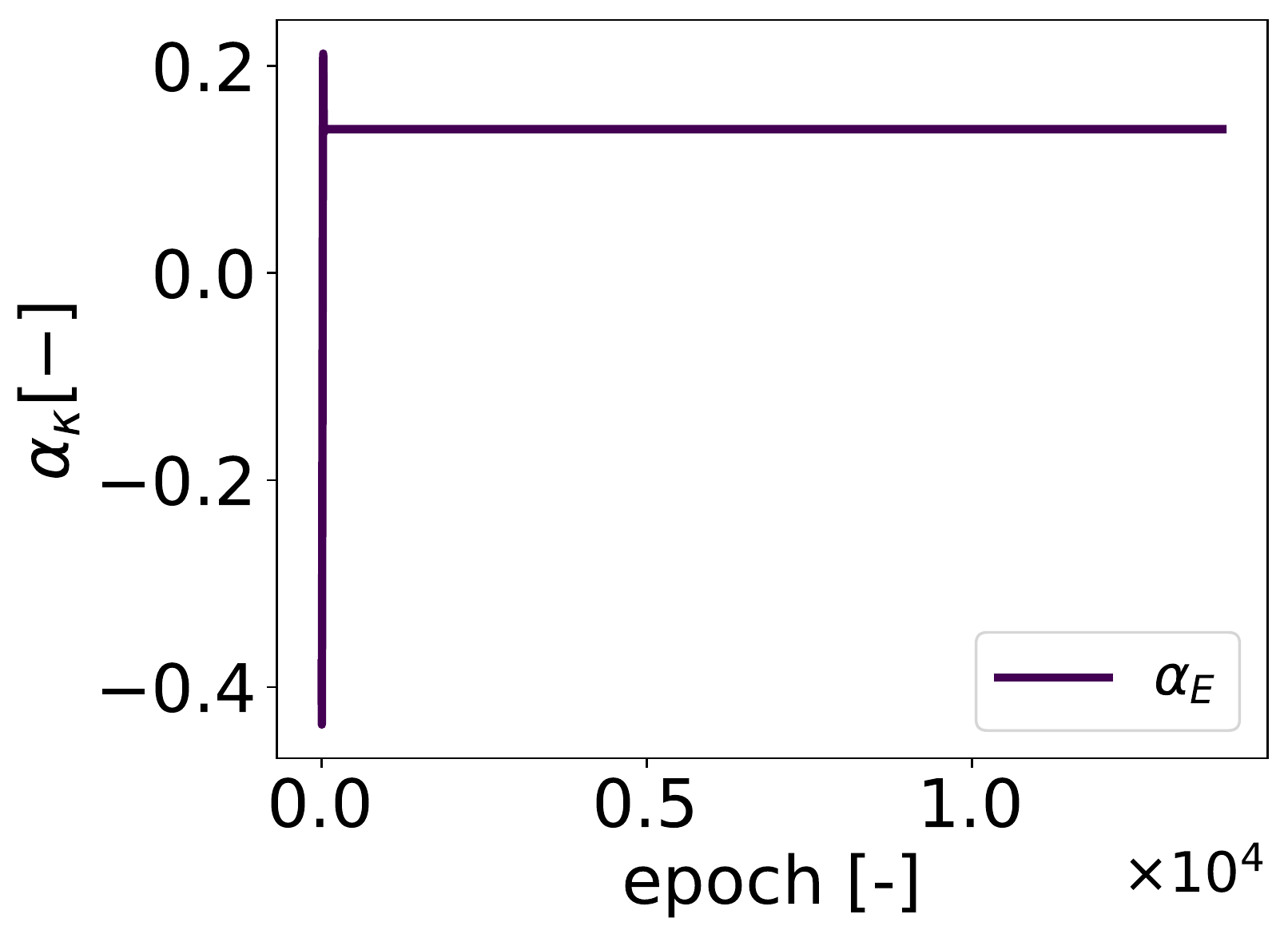}
        \caption{}
    \end{subfigure}
    \caption{\textbf{One-dimensional experimental data:} a) Displacement of tension rod approximated by a \ac{PINN} from experimental data and b) the evolution of the Young's modulus correction factor $\alpha_{E}$ during training. The displacement approximation is visually in good agreement with the measured data.}
    \label{fig:1D_experimental_displacement_parameter}
\end{figure}

The approximated displacement, which is shown in \cref{fig:1D_experimental_displacement_parameter}, is visually in good agreement with the measured displacements. The \ac{PINN} identified a Young's modulus of $E_{ident} = 239{,}195.53 \, \frac{\text{N}}{\text{mm}^{\text{2}}}$, which differs by $RE_{E} = 7.41 \%$ from the value determined by the \ac{LS-FEM} approach. We manually selected the data loss weight so that the data and \ac{PDE} loss terms are reasonably balanced, similar to what can be seen in \cref{fig:2D_synthetic_example_loss_parameters} for the full-field displacement data. We assume that the noise in the data is the main reason for the observed discrepancy between the Young's modulus determined using the enhanced \ac{PINN} and \ac{LS-FEM} approach. Therefore, we examine the influence of different noise levels on the identification in more detail in \cref{sec:2D_data}.

\section{Parameter identification from full-field displacement data}\label{sec:2D_data}
In the next step, we apply the enhanced \ac{PINN} formulation to synthetically generated displacement data of a plate with a hole, considering both clean and noisy data. As in the previous section, we first conduct an error sensitivity analysis for the identified Young's modulus and Poisson's ratio with respect to the initial estimate. Furthermore, we show how the relative errors of the identified material parameters evolve with an increasing number of collocation points. Finally, we investigate the error sensitivity of the identified material parameters with respect to the noise level of the displacement data.

\subsection{Clean synthetic displacement data}\label{subsec:2D_data_clean}
\begin{figure}[htb]
    \centering
    \begin{tikzpicture}
        \newcommand\XO{0}
        \newcommand\YO{0}
        \coordinate (A) at (\XO,\YO);
        \coordinate (B) at (\XO+4.5,\YO);
        \coordinate (C) at (\XO+5,\YO+0.5);
        \coordinate (D) at (\XO+5,\YO+5);
        \coordinate (E) at (\XO,\YO+5);
        \draw (A) -- (B);
        \draw (C) arc (90:180:0.5);
        \draw (C) -- (D);
        \draw (D) -- (E);
        \draw (E) -- (A);
        \draw[->, thick] (\XO,\YO+0) -- ++(180:1);
        \draw[->, thick] (\XO,\YO+1) -- ++(180:1);
        \draw[->, thick] (\XO,\YO+2) -- ++(180:1);
        \draw[->, thick] (\XO,\YO+3) -- ++(180:1);
        \draw[->, thick] (\XO,\YO+4) -- ++(180:1);
        \draw[->, thick] (\XO,\YO+5) -- ++(180:1);
        \draw[thick] (\XO-1,\YO) -- (\XO-1,\YO+5);
        \node (BC_bottom)[below] at (\XO+2.5,\YO-0.75) {$\bar{u}_{y}=0$};
        \node (BC_right)[right] at (\XO+5.75,\YO+2.5) {$\bar{u}_{x}=0$};
        \node (BC_left)[left] at (\XO-1.25,\YO+2.5) {$\bar{\bm{t}}=\begin{bmatrix}
            -100 \, \frac{N}{mm^{2}} \\
            0
        \end{bmatrix}$};
        \node (BC_top)[above] at (\XO+2.5,\YO+5.25) {$\bar{\bm{t}}=\bm{0}$};
        \node (BC_hole)[below right] at (\XO+4.75,\YO) {$\bar{\bm{t}}=\bm{0}$};
        \draw[<->] (\XO,\YO-0.6) -- (\XO+5,\YO-0.6) node[midway, above, ] {$L = 100 \, \text{mm}$};
        \draw[<->] (\XO+5.6,\YO) -- (\XO+5.6,\YO+5) node[midway, above, rotate=90 ] {$L = 100 \, \text{mm}$};
        \draw[->] (\XO+5,\YO) -- ++(135:0.5) node[left] {R = 10 mm};
        \node at (\XO+7,\YO+4.75){$D = 1 \, \text{mm}$};
        \draw[->] (2.5,2.5) -- ++(90:1) node[midway, left] {y};
        \draw[->] (2.5,2.5) -- ++(0:1) node[midway, below] {x};
    \end{tikzpicture} 
    \caption{Geometry and boundary conditions of the top left quadrant of a plate with a hole under uniaxial tension. The plate is assumed to be in plane stress condition. Body forces are neglected.}
    \label{fig:setup_2D_synthetic}
\end{figure}
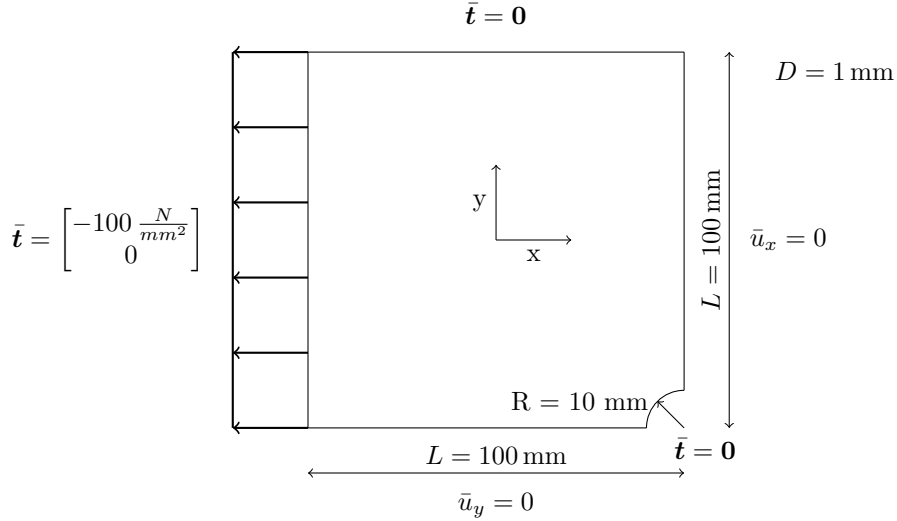

For the two-dimensional test case, we generate synthetic \ac{DIC} data for a plate with a hole using the \ac{FEM}. Since the plate with a hole is two-fold symmetric, we consider only the quadrant at the top left of the plate and define symmetry boundary conditions on the bottom and right boundaries. The quadrant has an edge length of $L=100 \, \text{mm}$ and the radius of the hole is $R = 10 \, \text{mm}$. Furthermore, the thickness of the plate is set to $T = 1 \, \text{mm}$ and the plate is assumed to be in plane stress condition. We simulate uniaxial tension and load the left edge with $\bar{\bm{t}} = [-100 \, \frac{\text{N}}{\text{mm}^{\text{2}}}, 0]^{T}$. No force is applied to the upper and the hole boundaries. External body forces are neglected. The geometry, dimensions and boundary conditions of the considered test case are illustrated in \cref{fig:setup_2D_synthetic}. We again assume isotropic, linear-elastic material. In two-dimensions, linear-elastic materials can be characterized by the Young's modulus and Poisson's ratio which are set to $E_{true} = 210{,}000 \, \frac{\text{N}}{\text{mm}^{\text{2}}}$ and $\nu_{true} = 0.3$, respectively, emulating the behaviour of steel. For the \ac{FE} simulation, the domain is meshed using linear triangular elements and the \ac{FE} solution is evaluated and recorded at a total of $101{,}496$ nodes. Discretization errors are neglected in the following due to the high resolution of the computational mesh. Under the given boundary conditions, the maximum calculated strain is $\varepsilon_{max} = 0.1505\%$ and thus in the assumed linear-elastic range of the considered material. The \ac{FEM} code for data generation is implemented in Python and based on the FEniCS project \cite{alnaes_FEniCSProjectVersion_2015}.

To solve the inverse problem, we only use a small subset of the synthetically generated displacement data consisting of $N_{o} = 4{,}096$ data points within the domain of interest. We select the data points directly from the \ac{FE} solution to avoid interpolation errors. In addition, the external work is approximated from $N_{W_{ext}} = 64$ equidistantly distributed data points on the left Neumann boundary. It should be noted that the displacement data set also contains points on the boundary. Therefore, we do not explicitly consider the Dirichlet boundary condition by a separate data set and loss term. The architecture of the \ac{PINN} and the training hyperparameters are defined as follows: We use two fully connected \acp{FFNN} to approximate the displacement fields $u_{x}(\bm{x})$ and $u_{y}(\bm{x})$. Both \ac{ANN} consist of $2$ input neurons, $2$ hidden layers with $16$ neurons each, and $1$ output neuron. The hyperbolic tangent is used as activation function in the hidden layers. The weights and biases of the \ac{ANN} are initialized using Glorot normal initialization and zeros, respectively. As in the previous test cases, we initialize the correction factors for bulk and shear modulus with zeros. We solve the resulting optimization problem using the BFGS optimizer.

In order to investigate the error sensitivities of the identified Young's modulus and Poisson's ratio with respect to their initial estimates, we again perform a sensitivity analysis. For this purpose, we consider initial estimates for both Young's modulus and Poisson's ratio of $66,\bar6 \%$, $100 \%$ and $133,\bar3 \%$ of the exact values used for data generation, respectively. Here we use the same number and locations of data and collocation points, so that $N_{o} = N_{col} = 4{,}096$. We then calibrate the linear-elastic material model from \cref{eq:constitutive_model_2D_K_G} for all of the resulting $9$ combinations of initial estimates using a \ac{PINN}. The \ac{ANN} parameters are initialized identically for each of the $9$ simulations of one analysis. To obtain more meaningful results, we again repeat the analysis a total of $10$ times with different random initializations of the \ac{PINN} parameters and locations of the data and collocation points.

\begin{figure}[htb]
    \centering
    \includegraphics[scale=0.65]{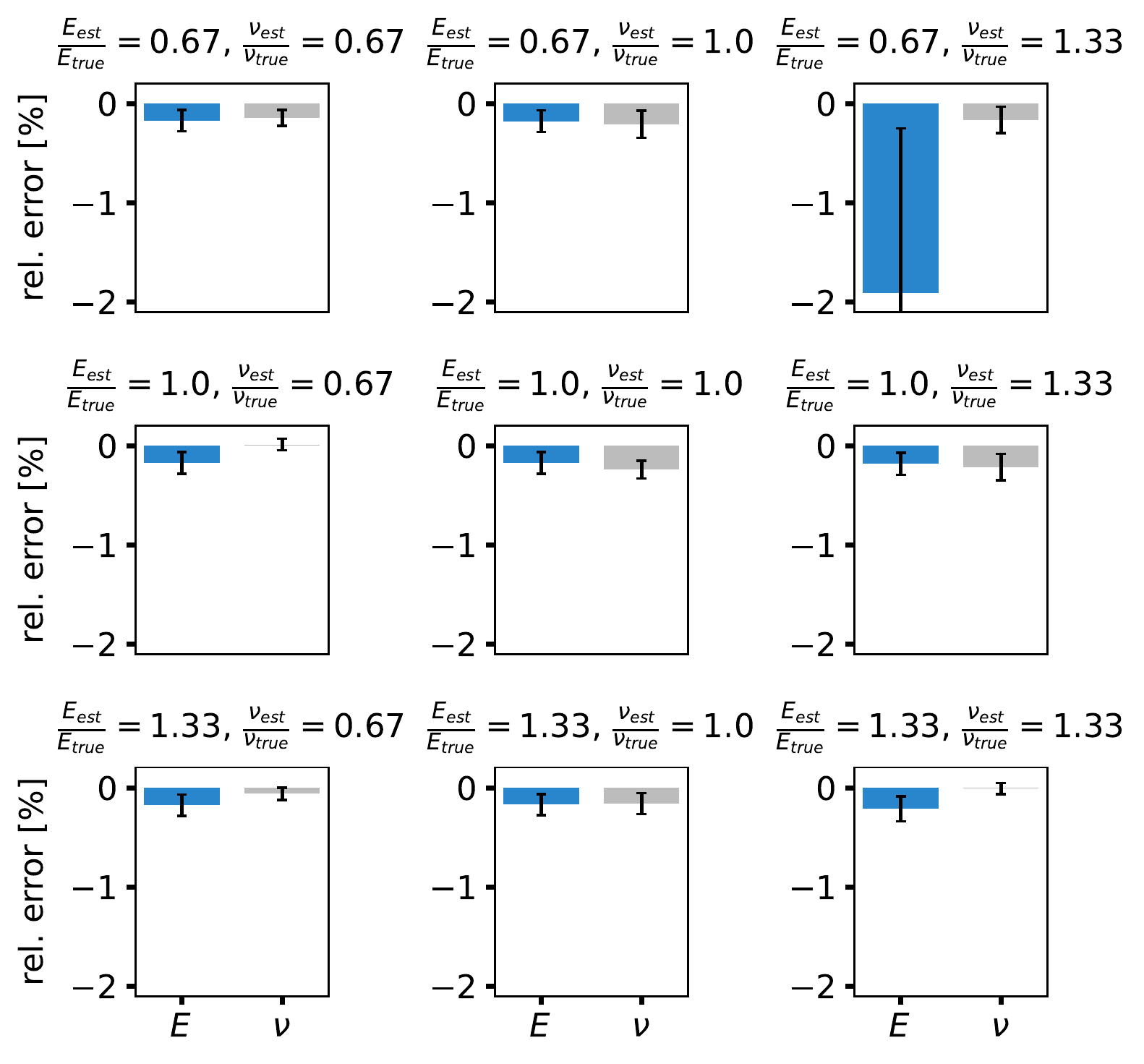}
    \caption{\textbf{Sensitivity analysis:} Error sensitivity of the identified Young's modulus $E$ and Poisson's ratio $\nu$ with respect to their initial estimates. We consider initial estimates for both Young's modulus and Poisson's ratio of $66,\bar6 \%$, $100 \%$ and $133,\bar3 \%$ of the exact values used for data generation, respectively. The simulation for each of the $9$ combinations of initial estimates is repeated a total of $10$ times with different random initializations of \ac{ANN} parameters and locations of the data and collocation points. The figure shows the mean and \acfp{SEM} of the relative errors calculated from the individual runs.}
    \label{fig:2D_synthetic_sensitivity_estimate}
\end{figure}

The results of the sensitivity analysis, summarized in \cref{fig:2D_synthetic_sensitivity_estimate}, show that the maximum \ac{MARE} of the identified material parameters averaged over all runs are $MARE^{max}_{E} = 1.91 \%$ and $MARE^{max}_{\nu} = 0.24 \%$ for Young's modulus and Poisson's ratio, respectively. If we consider all $90$ simulations separately, in the worst case, the maximum \acfp{ARE} are $ARE^{max}_{E} = 17.62 \%$ and $ARE^{max}_{\nu} = 1.46 \%$ for Young's modulus and Poisson's ratio, respectively. It should be noted that the above mentioned high $ARE^{max}_{E}$ also leads to the relatively high $MARE_{E}$ for $E_{est} = 140{,}000 \, \frac{\text{N}}{\text{mm}^{\text{2}}}$ and $\nu_{est} = 0.4$. For the remaining $9$ runs with this combination of initial estimates, the relative errors of the identified Young's modulus are less than $ARE_{E} = 0.9 \%$. Overall, the sensitivity analysis shows that the relative error has just a low sensitivity to the initial material parameter estimates within a reasonable range of estimates for real-world applications. Since the relative error for the identified Young's modulus was high in only $1$ out of $10$ runs, it is not a systematic error. Instead, the random initialization of the neural network parameters and the random distribution of the data and collocation points seem to have an impact on the accuracy.

\begin{figure}[htb]
    \centering
    \begin{subfigure}{0.49\textwidth}
        \centering
        \includegraphics[width=0.9\linewidth]{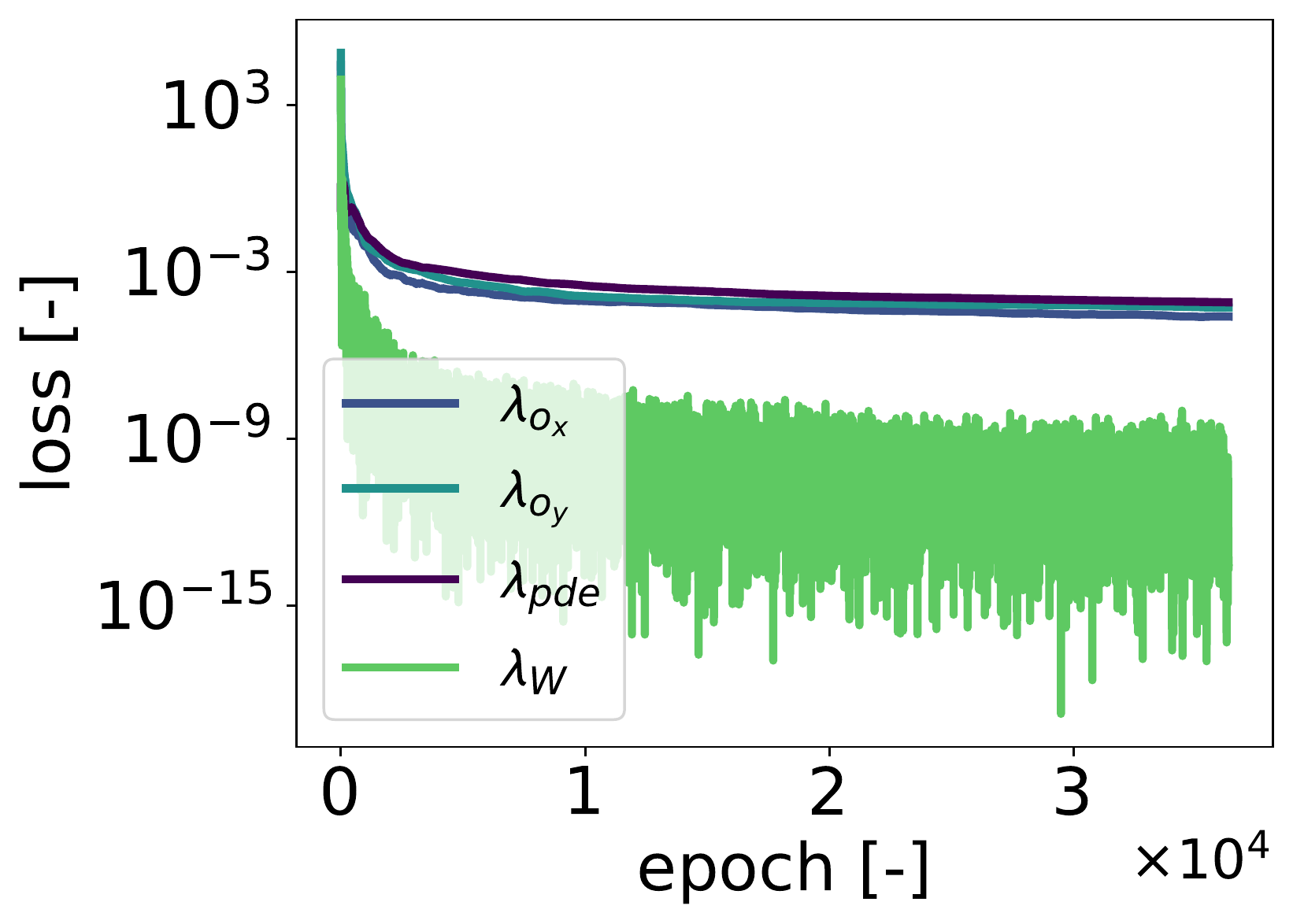}
        \caption{}
    \end{subfigure}
    \begin{subfigure}{0.49\textwidth}
        \centering
        \includegraphics[width=0.9\linewidth]{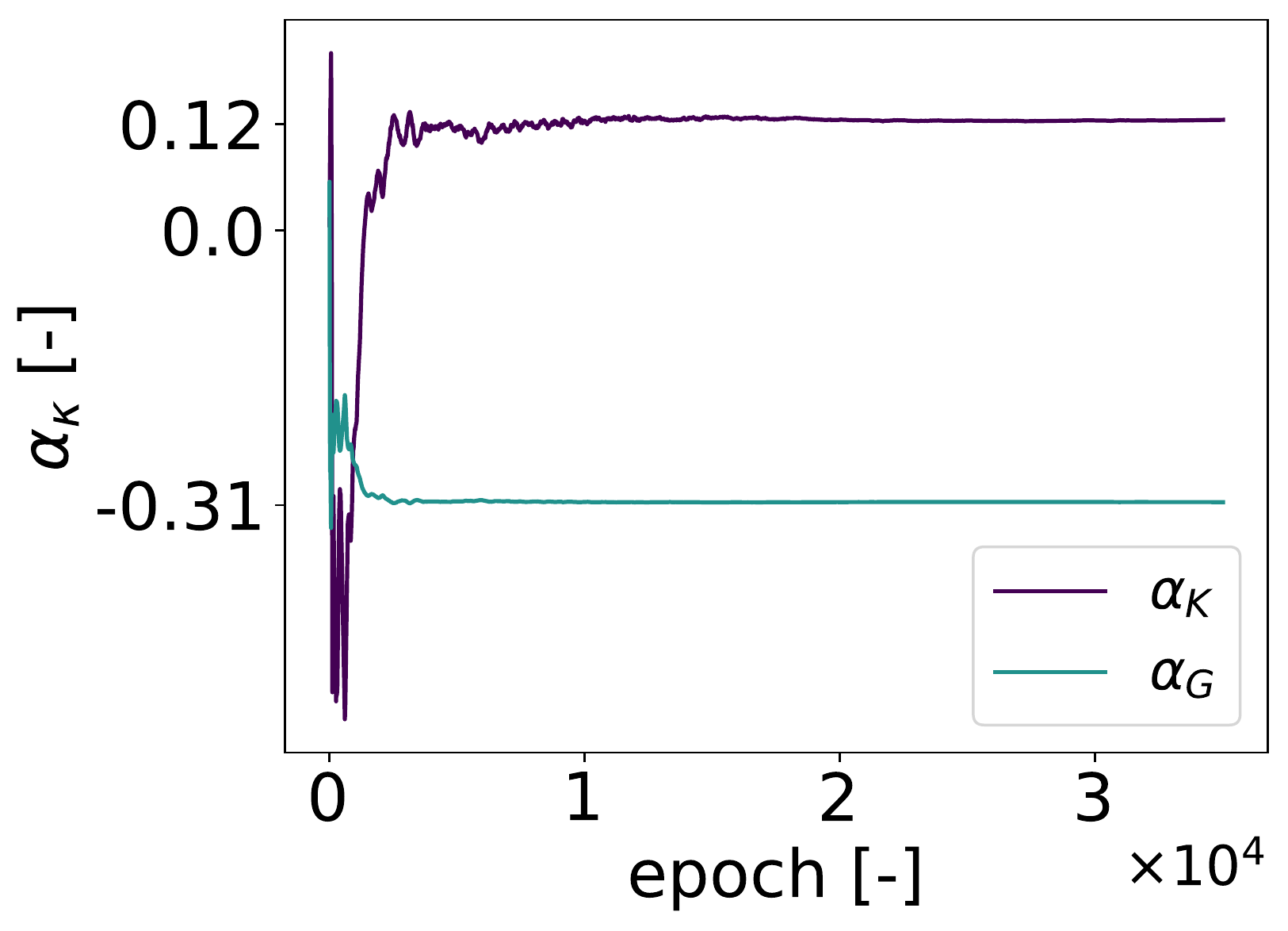}
        \caption{}
    \end{subfigure}
    \caption{\textbf{Clean two-dimensional data:} Exemplary evolution of a) the different loss terms and b) the correction factors $\alpha_{K}$ and $\alpha_{G}$. The correction factors relate to the initial estimates $E_{est} = 280{,}000 \, \frac{\text{N}}{\text{mm}^{\text{2}}}$ and $\nu_{est} = 0.2$ for Young's modulus and Poisson's ratio, respectively.}
    \label{fig:2D_synthetic_example_loss_parameters}
\end{figure}

\begin{figure}[htb]
    \centering
    \begin{subfigure}{0.49\textwidth}
        \centering
        \includegraphics[width=0.99\linewidth]{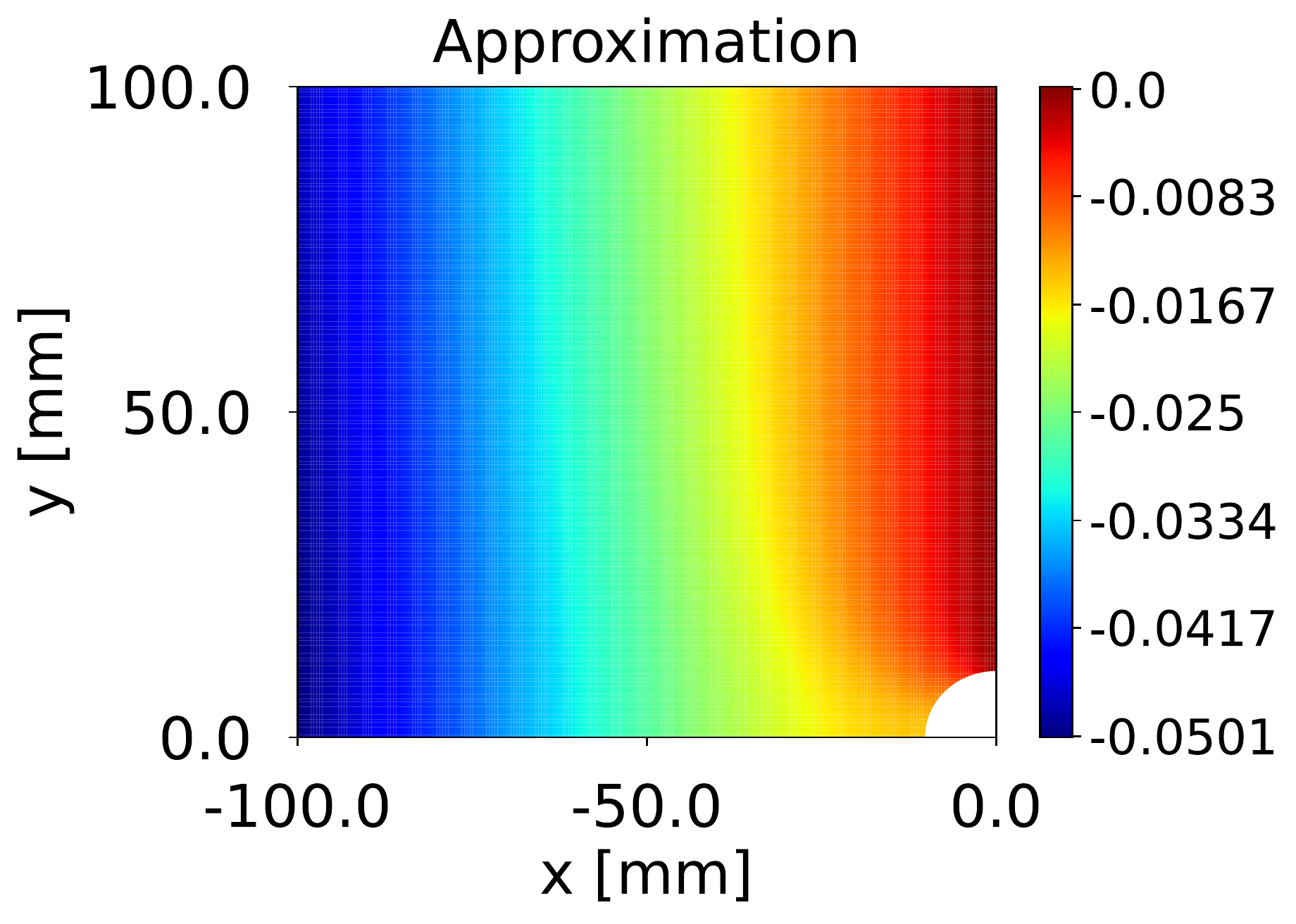}
        \caption{}
    \end{subfigure}
    \begin{subfigure}{0.49\textwidth}
        \centering
        \includegraphics[width=0.99\linewidth]{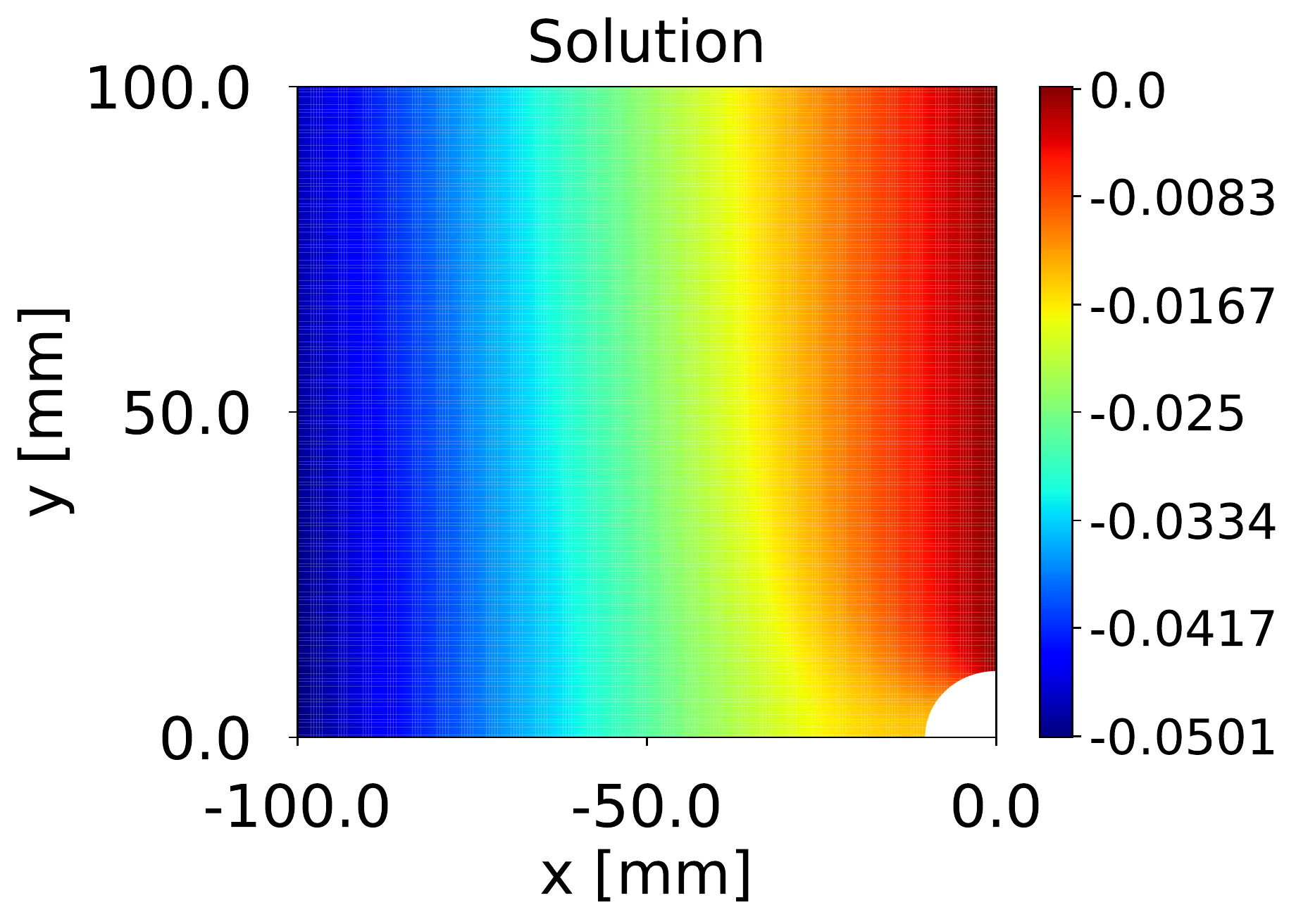}
        \caption{}
    \end{subfigure}
    \caption{\textbf{Clean two-dimensional data:} Exemplary illustration of a) the approximation and b) the solution of the displacement field in x-direction. The initial estimates for this example are $E_{est} = 280{,}000 \, \frac{\text{N}}{\text{mm}^{\text{2}}}$ and $\nu_{est} = 0.2$ for Young's modulus and Poisson's ratio, respectively. The displacements are measured in $mm$.}
    \label{fig:2D_synthetic_example_displacement}
\end{figure}

In \cref{fig:2D_synthetic_example_loss_parameters}, we show an example of how the loss terms and the correction factors for bulk and shear modulus for initial estimates of $E_{est} = 280{,}000 \, \frac{\text{N}}{\text{mm}^{\text{2}}}$  and $\nu_{est} = 0.2$ evolve during training. The identified bulk and shear modulus for this example result in \acfp{RE} of $RE_{E} = 0.15 \%$ and $RE_{\nu} = -0.11 \%$ for Young's modulus and Poisson's ratio, respectively. In figure \cref{fig:2D_synthetic_example_displacement}, we show the \ac{PINN} approximation of the displacement field after training in comparison with the \ac{FE} solution. To validate the accuracy of the displacement field, we use another $N_{val} = 4{,}096$ data points randomly sampled from the \ac{FE} solution and different from the training data. The displacement field is approximated with a relative $L^{2}\text{-norm}$ ($\text{r}L^{2}\text{-norm}$) of $\text{r}L^{2}_{x} = 1.1193 \cdot 10^{-5}$ in x- and $\text{r}L^{2}_{y} = 2.1224 \cdot 10^{-5}$ in y-direction. 

\begin{figure}[htb]
    \centering
    \includegraphics[scale=0.50 ]{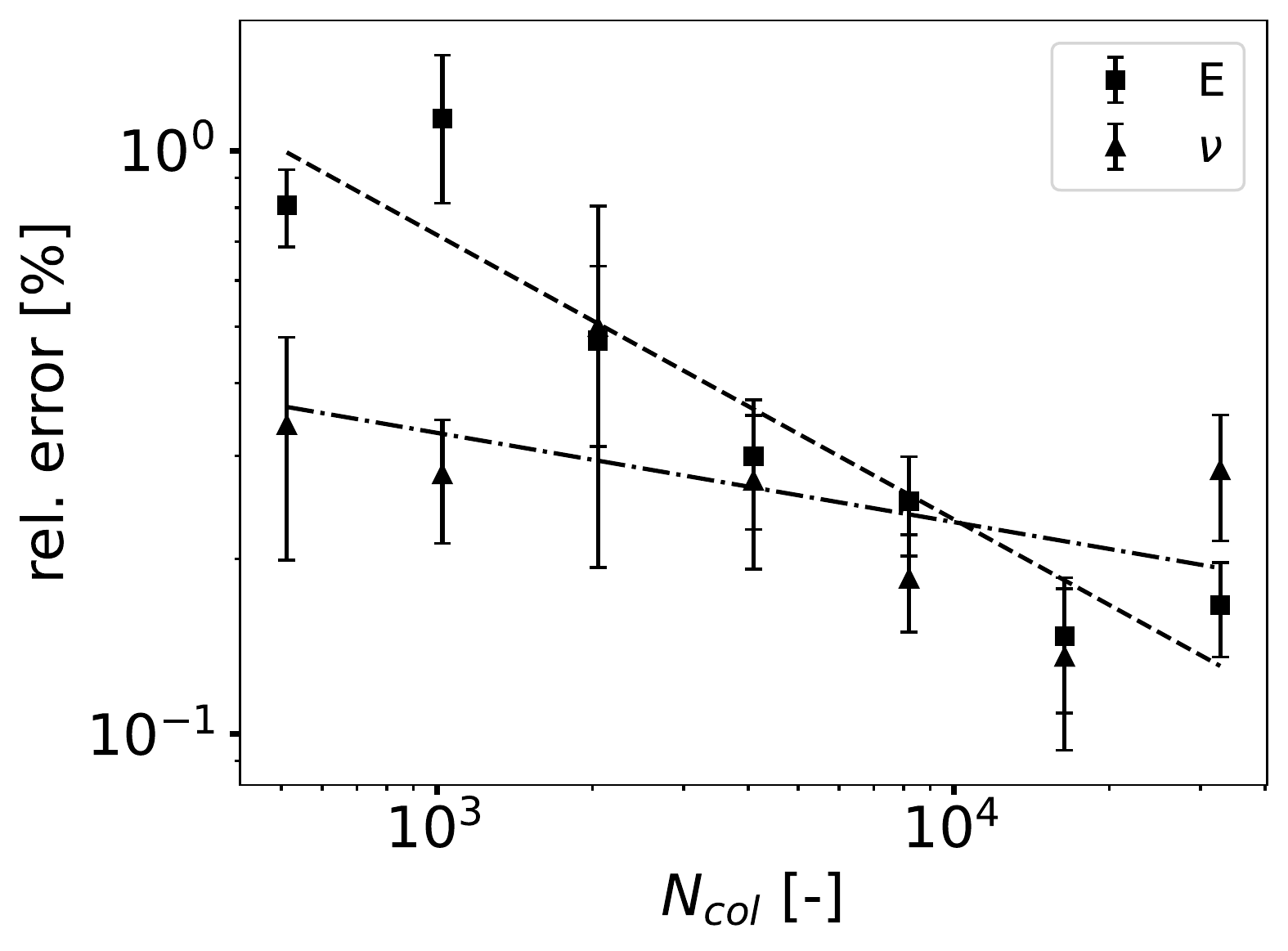}
    \caption{\textbf{Convergence study:} Relative errors of Young's modulus $E$ and Poisson's ratio $\nu$ with respect to the number of collocation points $N_{col}$ used to determine the \ac{PDE} residual and the internal work. The simulation for each tested number of collocation points is repeated a total of $10$ times with different random initializations of \ac{ANN} parameters and locations of the data and collocation points. The figure shows the mean and \acfp{SEM} of the relative errors calculated from the individual runs.}
    \label{fig:2D_synthetic_convergence}
\end{figure}

Note that the approximation of the integrals for calculating the internal and external work in \cref{eq:method_internal_work,eq:method_external_work} for heterogeneous strain and stress fields is erroneous for a finite number of collocation points. However, we assume that the accuracy of the approximation depends directly on the number of collocation points. To verify this assumption, we examine the convergence of the relative errors with respect to the number of  collocation points within the range $N_{col} = [512, \, 32{,}768]$. The number of data points $N_{o} = 4{,}096$ remains unchanged. We again repeat the convergence study a total of $10$ times with different randomly initialized \ac{ANN} parameters and locations for the data and collocation points to obtain more meaningful and robust results. For this analysis, we use the same \ac{PINN} architecture and hyperparameters as before and provide the material parameters we used for data generation as initial estimates. The results shown in \cref{fig:2D_synthetic_convergence} demonstrate that both the relative error of the identified Young's modulus and Poisson's ratio decrease with increasing number of collocation points. In direct comparison, the relative error for the identified Young's modulus decreases much faster.

\subsection{Noisy synthetic displacement data}\label{subsec:2D_data_noisy}
Finally, we investigate how sensitive the relative errors of the identified Young's modulus and Poisson's ratio are with respect to the level of noise. So far, we have not yet taken into account the fact that displacement data measured by a \ac{DIC} system is inevitably affected by noise. In order to emulate real \ac{DIC} data, we apply Gaussian noise $\mathcal{N}(0,\sigma^{2})$ with zero mean to the synthetic data used in \cref{subsec:2D_data_clean}. The noise in the \ac{DIC} measurements and thus also the standard deviation depend on the pixel accuracy of the imaging device and is not proportional to the displacement magnitude. For this reason, we have determined the standard deviation relative to the order of magnitude of the mean displacement, which is $10^{-2} \, \text{mm}$. We applied the same absolute noise level to all nodal values of the \ac{FEM} solution independently of the individual magnitude. For the sensitivity analysis, we apply varying levels of Gaussian noise $\mathcal{N}(0,\sigma^{2})$ to the clean data we used in the previous section in a range of $[0.1 \%, 10 \%]$ relative to the order of magnitude of the mean displacement. This results in absolute standard deviations in the range of $\sigma = [10^{-5}, 10^{-3}] \, \text{mm}$. Just as with the sensitivity analysis in \cref{subsec:2D_data_clean}, we use $N_{o} = N_{col} = 4{,}096$ training and collocation points. As initial estimates, we provide the same material parameters that we used to generate the clean displacement data. To avoid overfitting of the \acp{ANN} to the noisy data, we reduce the data loss weight in \cref{eq:method_summary_loss_function_2D} to $\lambda_{o} = 10^{3}$. Apart from that, we use the same \ac{PINN} architecture and hyperparameters as in \cref{subsec:2D_data_clean}.

\begin{figure}[htb]
    \centering
    \includegraphics[scale=0.55]{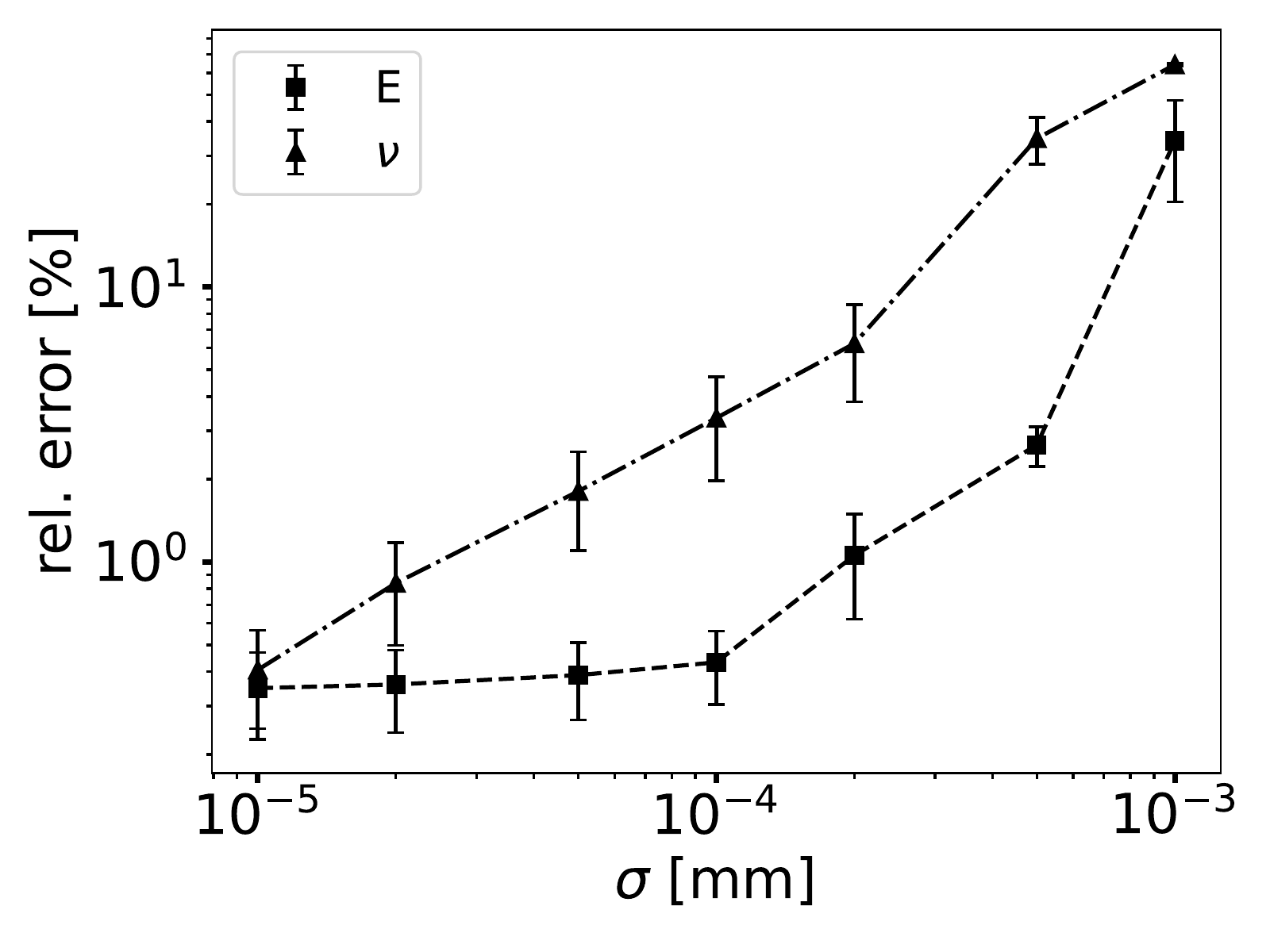}
    \caption{\textbf{Sensitivity analysis:} Relative errors of the identified Young's modulus $E$ and Poisson's ratio $\nu$ with respect to the level of noise. We assume Gaussian noise with zero mean and test various standard deviations in the range $\sigma = [10^{-5} \, \text{mm}, 10^{-3} \, \text{mm}]$. The simulation for each tested standard deviation is repeated a total of $10$ times with different random initializations of \ac{ANN} parameters and locations of the data and collocation points. The figure shows the mean and \acfp{SEM} of the relative errors calculated from the individual runs.}
    \label{fig:2D_synthetic_sensitivity_noise}
\end{figure}

The results of the sensitivity analysis, shown in \cref{fig:2D_synthetic_sensitivity_noise}, indicate that \acp{PINN} can deal with moderate levels of noise when calibrating the linear-elastic material model. From the results, it can be seen that especially the relative error of the identified Young's modulus increases only slightly in its absolute values up to a standard deviation of $\sigma = 10^{-4} \, \text{mm}$. The \acfp{MARE} at this level of noise are $MARE_{E} = 0.43 \%$ and $MARE_{\nu} = 3.34 \%$ for Young's modulus and Poisson's ratio, respectively. For larger standard deviations, the relative errors increase significantly. According to \cite{pierron_ExtensionVirtualFields_2010}, a representative value for the standard deviation of Gaussian random noise in \ac{DIC} measurements is $\sigma = 4 \cdot 10^{-4} \, \text{mm}$. However, this requires optimal conditions of the experiment, which includes, among others, the optical setup as well as the imaging device. To account for the fact that these are not always met in practice, the authors test the \ac{PINN} for a standard deviation of $\sigma = 5 \cdot 10^{-4} \, \text{mm}$. Under these assumptions, the mean absolute relative errors for a realistic standard deviation of the noise in \ac{DIC} measurements are $MARE_{E} = 2.67 \%$ and $MARE_{\nu} = 34.68 \%$ for Young's modulus and Poisson's ratio, respectively. In particular, for the Poisson's ratio, the relative error is no longer in an acceptable range. In future work, we therefore aim to further enhance the method to increase the methods performance in the presence of noisy data as we discuss in the next section.

\section{Conclusion and outlook}\label{sec:conclusion}
\Acp{PINN} have emerged as a suitable alternative approach to traditional numerical methods, such as \ac{LS-FEM} or \ac{VFM}, to solve inverse problems in the field of solid mechanics. It has recently been shown that \acp{PINN} are in principal capable of identifying the material parameters of constitutive models from full-field displacement data \cite{zhang_AnalysisInternalStructures_2022,haghighat_PhysicsinformedDeepLearning_2021,hamel_CalibratingConstitutiveModels_2022,zhang_PhysicsInformedNeuralNetworks_2020}. However, the assumptions made in the literature often do not match real-world conditions. Indeed, as soon as the standard \ac{PINN} is applied to realistic data, problems arise that have so far been neglected in the literature. In addition, not all approaches in the literature realize the full potential of \acp{PINN}. This potential lies in the fact that the inclusion of observation data is straightforward and no computational grid is required.

In this contribution, we have done the first steps to further enhance the standard \ac{PINN}, as proposed in \cite{raissi_PhysicsinformedNeuralNetworks_2019}, towards full-field displacement data in a realistic regime. The realistic data regime refers to a realistic order of magnitude of the displacement data and material parameters as well as the fact that the data is inevitably affected by noise. The focus of our contribution was on the conditioning and reformulation of the resulting optimization problem. In this process, we first normalized the input and output of the \ac{PINN} and scaled the parameters to be optimized by providing initial estimates for the material parameters. A sensitivity analysis then demonstrated that the relative errors of the identified material parameters have only a low sensitivity to the initial estimates in a reasonable range for real-world applications. Since we found that the cost function was ill-conditioned for realistic displacement data, we balanced the latter by weighting the data loss term. In order to reduce the dependence of the identified material parameters on local approximation errors in the boundary region, we based the identification not on the stress boundary condition but instead on the global balance of internal and external work. In a convergence study, we were able to show that the errors of the identified material parameters converge with the number of collocation points used to approximate the integral for calculating the internal work and the \ac{PDE} residual. Furthermore, we found that the resulting optimization problem is better posed when it is formulated in terms of bulk and shear modulus. Finally, we investigated the sensitivity of the identified Young's modulus and Poisson's ratio with respect to various levels of noise. 

Although the enhanced \ac{PINN} shows satisfactory results for clean as well as moderately noisy data, future work is still needed for a real-world application of \acp{PINN}, e.g., as part of a monitoring system as motivated in the introduction. Future work includes, but is not limited to, increasing the robustness of the method to noise in the measurement data. It needs to be investigated whether this can be achieved by quantifying and accounting for the uncertainties in the identified material parameters. For this purpose, we will consider extended variants of \acp{PINN} \cite{zhang_QuantifyingTotalUncertainty_2019,yang_BPINNsBayesianPhysicsinformed_2021}. Since the data loss term is currently weighted manually, we also plan to investigate methods for adaptively balance the cost function, see e.g. \cite{wang_UnderstandingMitigatingGradient_2021, wang_WhenWhyPINNs_2022}. A drawback of the proposed method is that the \ac{PINN} must be re-trained for each full-field displacement measurement. To accelerate the convergence of the identification process, future work should investigate and validate different approaches, such as adaptive activation functions \cite{jagtap_AdaptiveActivationFunctions_2020}. Furthermore, the results show that the accuracy of the calibration depends on the initialization of the ANN parameters and the location of the data and collocation points. To improve the robustness of the method, this random factor must be resolved. In order to be able to evaluate the quality of the identified material parameters in real-world applications, qualitative metrics, such as derived in \cite{hartmann_IdentifiabilityMaterialParameters_2018} in the context of the \ac{LS-FEM} approach, are required.

Another challenge in real-world applications is to take into account material inhomogeneities and to detect fracture at an early stage. Therefore, we intend to further develop the method towards heterogeneous materials and approximate the material field by another \ac{ANN}, as proposed in \cite{zhang_PhysicsInformedNeuralNetworks_2020}. The determined parameter field could then also be the basis for identifying fracture in the material. In this paper, using the linear-elastic material model, we have pointed out some issues that arise for a realistic data regime and have conditioned the optimization problem. In principle, the first four extensions presented in \cref{sec:method_development} are not restricted to linear problems. The method presented here can therefore also be applied to more complex material models in combination with more sophisticated mechanical test cases. Basically, the method presented in this contribution, which follows the all-at-once approach, is not the only option for material model calibration using \acp{PINN}. A concurrent approach is based on parametric \acp{PINN} \cite{beltranpulido_ParametricPINNs_2022}, where the \ac{PINN} acts as a surrogate and learns the parameterized solution of the underlying parametric \ac{PDE}. The pre-trained surrogate can then be used for model calibration. From the authors' perspective, therefore, conceptual work is prioritized before applications to more complex materials and geometries will be considered.


\section*{Declaration of competing interests} \noindent
The authors declare that they have no known competing financial interests or personal relationships that could
have appeared to influence the work reported in this paper.

\section*{Acknowledgement} \noindent
The support of the Deutsche Forschungsgemeinschaft (DFG, German Research Foundation) is gratefully acknowledged in the following projects:
\begin{itemize}
    \item DFG 255042459/GRK2075-2: \textit{Modelling the constitutional evolution of building materials and structures with respect to aging}.
    \item DFG 501798687: \textit{Monitoring data driven life cycle management with AR based on adaptive, AI-supported corrosion prediction for reinforced concrete structures under combined impacts}. Subproject of SPP 2388: \textit{Hundred plus - Extending the Lifetime of Complex Engineering Structures through Intelligent Digitalization}.
\end{itemize}
In both projects, the solution of inverse problems is a key enabler to link measurement data and physical models. We further thank the Institute of Applied Mechanics at Technische Universität Clausthal for providing us the experimental data and the reference results of the \ac{LS-FEM} approach. The first author wants also to thank Alexander Henkes, Jendrik-Alexander Tröger, Knut Andreas Meyer and Ralf Jänicke for the fruitful discussions.

\section*{Data availability} \noindent
The code will be published on zenodo.org upon acceptance of this manuscript.

\bibliography{literature}
\bibliographystyle{elsarticle-num}

\end{document}